\documentclass[11pt]{article}

\PassOptionsToPackage{hyphens}{url}
\usepackage[preprint]{acl}
\usepackage{times}
\usepackage{latexsym}
\usepackage[T1]{fontenc}
\usepackage[utf8]{inputenc}
\usepackage{inconsolata}
\usepackage{url}
\usepackage{graphicx}
\usepackage{booktabs}
\usepackage{multirow}
\usepackage{amsmath}
\usepackage{amssymb}
\usepackage{array}
\newcommand{\sopivottablestyle}{%
  \small
  \renewcommand{\arraystretch}{1.12}%
  \setlength{\tabcolsep}{4pt}%
}

\usepackage{placeins}
\usepackage{stfloats}
\usepackage{microtype}
\usepackage{listings}
\usepackage{algorithm}
\usepackage{algpseudocode}
\usepackage[breakable,listings,skins]{tcolorbox}
\lstdefinestyle{sopivotprompt}{basicstyle=\small\rmfamily,
  breaklines=true,breakatwhitespace=false,breakautoindent=false,breakindent=0pt,columns=fullflexible,
  keepspaces=true,showstringspaces=false,frame=none,
  literate={_}{{\_\allowbreak}}1,
  aboveskip=0pt,belowskip=0pt,
  morekeywords={SYSTEM,USER},keywordstyle=\bfseries}
\newcounter{sopivotprompt}
\newcommand{\sopivotpromptbox}[2]{%
  \par%
  \begin{nolinenumbers}%
  \refstepcounter{sopivotprompt}%
  \tcbinputlisting{enhanced,breakable,lines before break=10,listing only,listing engine=listings,
    listing file={#2},listing options={style=sopivotprompt},
    title={Prompt \thesopivotprompt: #1},
    title after break={Prompt \thesopivotprompt\ (continued)},
    colback=white,colbacktitle=black!7,coltitle=black,
    colframe=black!35,boxrule=0.3pt,arc=0pt,outer arc=0pt,
    fonttitle=\small\bfseries,left=6pt,right=6pt,top=5pt,bottom=5pt,
    toptitle=3pt,bottomtitle=3pt,before skip=6pt,after skip=8pt}%
  \end{nolinenumbers}}
\floatstyle{plain}
\newfloat{promptpanel}{p}{lop}
\floatname{promptpanel}{Prompt}
\newcommand{\sopivotwidepromptpart}[5]{%
  \begin{promptpanel*}[#5]
  \renewcommand{\thepromptpanel}{#1}
  \begin{nolinenumbers}
  \tcbinputlisting{enhanced,listing only,listing engine=listings,
    listing file={#3},listing options={style=sopivotprompt},
    title={Prompt #1: #2},
    colback=white,colbacktitle=black!7,coltitle=black,
    colframe=black!35,boxrule=0.3pt,arc=0pt,outer arc=0pt,
    fonttitle=\small\bfseries,left=6pt,right=6pt,top=5pt,bottom=5pt,
    toptitle=3pt,bottomtitle=3pt,before skip=0pt,after skip=0pt}
  \end{nolinenumbers}
  \caption{#4}
  \end{promptpanel*}}
\usepackage{hyperref}
\title{Agentic Social Strategy Discovery: Learning What Matters from Branched Futures}
\author{
 \textbf{Linxiao Gong\textsuperscript{1,*}},
 \textbf{Tianfu Wang\textsuperscript{1,*}},
 \textbf{Songyang Gao\textsuperscript{1,*}},
 \textbf{Yin Wu\textsuperscript{1}},
 \textbf{Zhezheng Hao\textsuperscript{2}},
\\
 \textbf{Yijie Xu\textsuperscript{1}},
 \textbf{Yichen Wan\textsuperscript{3}},
 \textbf{Liwei Deng\textsuperscript{4}},
 \textbf{Yutao Yue\textsuperscript{1}},
 \textbf{Hui Xiong\textsuperscript{5,\dag}}
\\
\\
 \textsuperscript{1}The Hong Kong University of Science and Technology (Guangzhou)\\
 \textsuperscript{2}Zhejiang University\quad
 \textsuperscript{3}Baidu\quad
 \textsuperscript{4}Aalborg University\\
 \textsuperscript{5}The Hong Kong University of Science and Technology\\
 \textsuperscript{*}Equal contribution.\quad
 \textsuperscript{\dag}Corresponding author: Hui Xiong.\\
 \texttt{xionghui@ust.hk}
}

\begin{document}

\maketitle

\begin{abstract}
Social intelligence requires language-model agents not only to engage in
multi-turn interactions but also to learn reusable strategies from past
experience. However, overall interaction scores offer limited insight into
which strategy choices change subsequent outcomes in a given context.
We present SoPivot, a framework for discovering social strategies supported
by local comparisons of alternative dialogue futures under the same decision
context. SoPivot samples repeated natural continuations for candidate actions,
selects strategies based on terminal outcomes, and distills them into situated
memories grounded in visible context. General strategies abstracted from
these memories undergo source-context replay screening; both memory levels
are retrieved for selective reuse in new interactions. On SOTOPIA-All,
SoPivot improves goal achievement over no memory by 0.596 and 0.481 points
under two model judges, and achieves the highest seven-dimension
average score across all four combinations of judge and evaluation subset
(All and Hard). On AgentSense, the transferred memories achieve a Goal score
of 89.00\%, exceeding no memory and MaTTS by 1.95 and 1.40 percentage points,
respectively. These results support distilling local interaction outcomes
into reusable strategies as an effective approach to improving language
models' social interaction performance.
\end{abstract}

\section{Introduction}

Language agents increasingly negotiate, collaborate, and resolve conflicts
with people in multi-turn interactions
\citep{zhou2024sotopia,mou2025agentsense}. Social intelligence in these settings
is inherently contextual and sequential. Agents must interpret others'
behavior, adapt to evolving social context, and anticipate how current choices
may affect later responses. Together, these capabilities support social
strategies, which are context-dependent approaches to communicating,
coordinating, and negotiating with others. Because different strategies can
redirect the same interaction toward different downstream outcomes, selecting
an appropriate strategy at a consequential moment is central to effective
social behavior. Learning from interaction experience therefore requires
identifying which strategy choices matter at consequential decision points,
rather than merely remembering whether an entire dialogue succeeded.

\begin{figure}[t]
\centering
\includegraphics[width=\columnwidth]{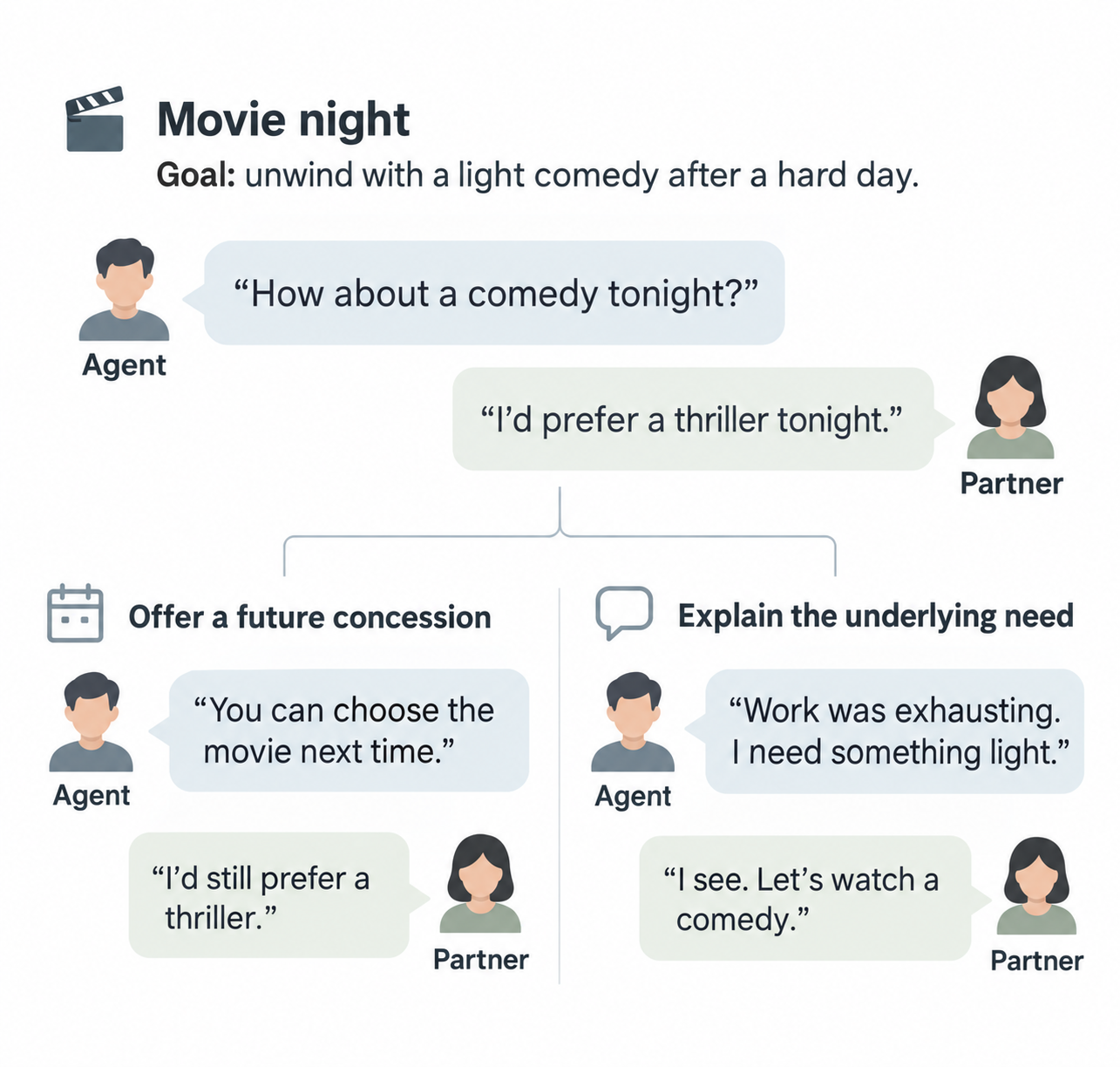}
\caption{Two illustrative continuations from the same movie-choice
disagreement. Offering a future concession and explaining the underlying
need elicit different replies.}
\label{fig:motivation}
\end{figure}

Existing approaches improve social interaction through policy learning,
strategy planning, and experience reuse. Prior work also derives
reusable memory from contrasting experiences: PRINCIPLES combines successful
revisions with failed trials \citep{kim-etal-2025-principles}, while parallel
MaTTS contrasts multiple trajectories for the same query
\citep{ouyang2026reasoningbank}. We study how such evidence can be organized
around a local social decision and its downstream consequences, since
immediate progress need not imply eventual goal completion, and complete
trajectories can differ at many decisions at once. SoPivot instead fixes the
pre-action context and a set of alternative actions, samples repeated natural
continuations for each, and compares terminal goal outcomes before
extracting situated strategies.

We therefore seek to make \emph{pivotal social strategies} explicit objects of
discovery. A pivotal social strategy is a context-dependent strategy choice
that redirects downstream interaction outcomes relative to plausible
alternatives available at the same decision point. Our evidence unit for identifying such strategies is a sampled comparison of alternative actions at a fixed decision context: each candidate is evaluated through repeated natural continuations from the same pre-action state, which gives a finer-grained unit of analysis than a whole trajectory and separates same-context support from later reuse evaluation. Grounding a
contrast-supported candidate in its visible context yields a situated
strategy that further specifies when it may apply. As illustrated in
Figure~\ref{fig:motivation}, offering a future concession and explaining the
underlying need are both plausible responses to the same movie-choice
disagreement, but they open different dialogue futures. This example motivates same-context comparison: repeated branch outcomes provide
evidence about a local strategy choice, whereas the success of either
trajectory in isolation provides no such evidence.

Here, we introduce \textbf{SoPivot}, an agentic framework for discovering
pivotal social strategies from branched futures.\footnote{Code and accompanying research materials will be publicly released.} After a
simulated source interaction is complete, SoPivot identifies candidate
critical decision points, constructs functionally distinct sibling actions
under the same visible decision context, and samples repeated natural
continuations after each action. Branch Outcome Comparison then identifies
contrast-supported strategies while retaining all branch outcomes in the
audit record. SoPivot grounds each contrast-supported strategy in visible
context to form a situated strategy, subject to extraction and
source-fidelity checks, and abstracts compatible situated entries into
general strategies, which are admitted when their pooled mean Goal in
repeated source-episode replays matches or exceeds the no-memory control.
For a new interaction, the system retrieves multiple optional strategies
once from the frozen situated and general memory. On SOTOPIA, we compare
SoPivot with no-memory and experience-memory baselines; component ablations
examine the role of source outcome evidence and the two memory levels. A
second model judge tests sensitivity to evaluation, and AgentSense measures
reuse on another interaction benchmark.

We make three contributions.
\begin{itemize}
    \item We formulate social strategy discovery around local action
    alternatives and their downstream goal outcomes, using repeated
    continuations from a shared pre-action context.
    \item We introduce SoPivot to identify candidate critical decision points
    and obtain contrast-supported strategies through Branch Outcome Comparison.
    \item We turn local contrast evidence into situated and general strategy
    memory, screen general strategies through source replay, and evaluate
    goal completion, memory components, and cross-benchmark reuse.
\end{itemize}

\section{Related Work}

\subsection{Social Intelligence in LLMs}

General-purpose LLMs, including recent foundation models such as Kimi K3
\citep{team2026kimi}, support two complementary routes for studying social
capabilities. The first evaluates social behavior through interactive
benchmarks such as SOTOPIA and AgentSense
\citep{zhou2024sotopia,mou2025agentsense}, alongside related work on
negotiation \citep{lewis2017deal,chawla2021casino}, emotional support
\citep{liu2021esc}, goal-oriented tutoring \citep{wang2025llm}, social-skill
practice \citep{wang2026socialcoach}, personalized human simulation
\citep{lei2026humanllm}, and large-scale social simulation
\citep{park2023generative}. The second improves social policies through
behavior cloning, self-reinforcement, strategy injection, and reinforcement
learning \citep{wang2024sotopiapi,zhang2025sotopiaomega,yu2025sotopiarl}.
Together, these routes provide environments and learning procedures for
social interaction. SoPivot focuses on a complementary unit: reusable
strategy knowledge grounded in comparisons of local actions and their
subsequent outcomes. It stores this knowledge in external memory for later
interactions without updating the acting model's weights.

\subsection{Experience-Driven Agent Memory}

Experience-driven memory follows two main routes: extracting reusable
knowledge from experience, and validating or maintaining that knowledge.
The first includes verbal feedback and cross-task insights
\citep{shinn2023reflexion,zhao2024expel}, reusable workflows and executable
skills \citep{wang2025awm,wang2024voyager}, and strategies induced from expert
dialogues \citep{yang2026metro}. Several closely related methods explicitly
use contrasting experiences. PRINCIPLES revisits the state before a failed
turn, revises the strategy and action, and combines successful revisions with
failed attempts to derive principles for later retrieval and reinterpretation
\citep{kim-etal-2025-principles}. Parallel MaTTS jointly contrasts successful
and failed complete trajectories for the same query
\citep{ouyang2026reasoningbank}. SkillCAT extracts skill patches from
same-task success/failure pairs and screens them through source-task replay
before merging \citep{chen2026skillcat}.

The second route tests and maintains stored knowledge. SkillAudit compares
executions with and without a skill to guide revision
\citep{gao2026skillaudit}; other systems study library management, held-out
validation and invalidation checks
\citep{pu2026skillops,yang2026skillopt,chao2026stale}. A parallel benchmark
line evaluates memory use across social episodes
\citep{goel2025lifelongsotopia}. Rollout-based methods also exploit alternative
futures and delayed outcomes, primarily for immediate action selection or
policy learning \citep{zhou2023lats,zhang2026mica,shi2026r3l}.

SoPivot connects local outcome comparison with reusable strategy memory.
Its comparison unit is a fixed set of alternative actions at a shared
pre-action context, each followed by repeated natural continuations to
terminal outcomes, which differs from iterative failure revision in
PRINCIPLES and complete-trajectory contrast in MaTTS. The resulting evidence
supports situated strategy extraction; compatible entries are then abstracted
into general strategies and screened through source replay. Reusable text
and retrieval remain restricted to information visible to the focal agent.

\section{Methodology}
\label{sec:method}

\begin{figure*}[t]
\centering
\includegraphics[width=\textwidth]{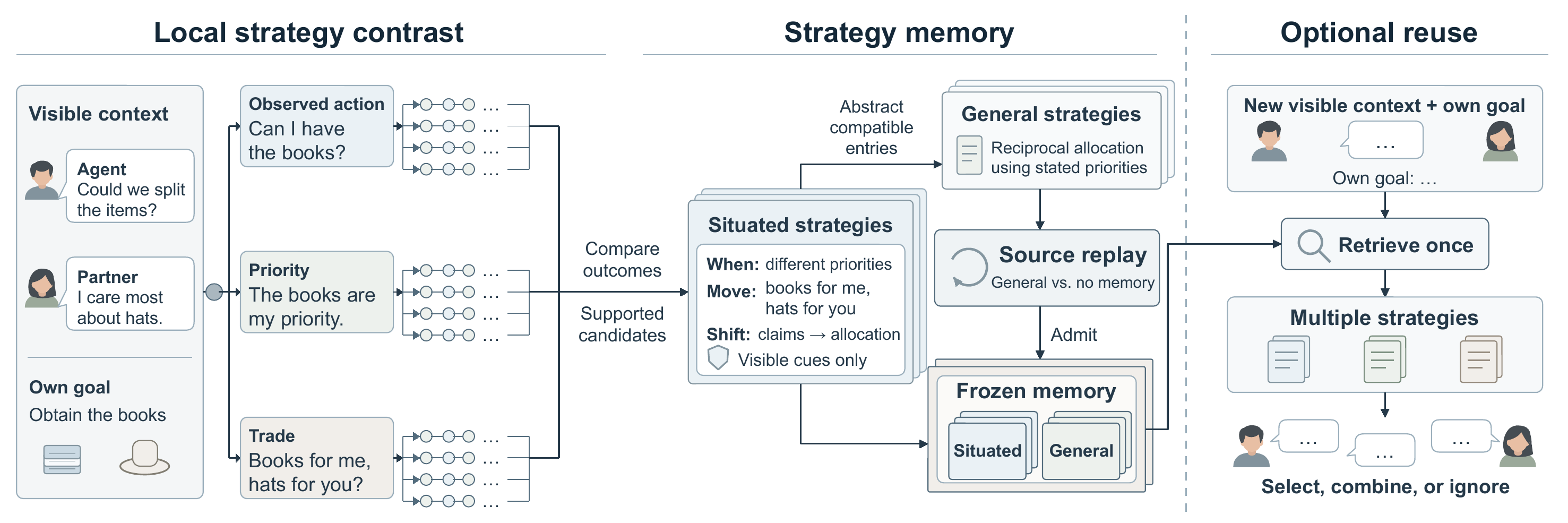}
\caption{SoPivot pipeline: local sibling comparison, strategy memory construction,
and optional reuse in new interactions. Memory uses only agent-visible information.}
\label{fig:pipeline-example}
\end{figure*}

Figure~\ref{fig:pipeline-example} gives an overview of SoPivot: local
strategy contrast (left), strategy memory construction (middle), and optional
reuse (right). At selected candidate critical decision points, SoPivot
constructs sibling strategy branches and compares their outcomes under the
same visible context. It then extracts situated strategies from
contrast-supported candidates under the decision-time information boundary,
abstracts related cases into general strategies, and replays each
abstraction on its sources. At reuse time, one initial query retrieves
multiple strategies from both memory levels. Appendix~\ref{app:method-details}
details the procedures; Appendices~\ref{app:algorithms} and~\ref{app:prompts}
give pseudocode and prompt templates.

\subsection{Problem Formulation}
\label{sec:problem}

Following the interaction setting introduced by SOTOPIA and adopted in
subsequent work \citep{zhou2024sotopia,yu2025sotopiarl}, we consider an agent
engaged in a goal-oriented, multi-turn social interaction with another agent.
Let $x$ denote the scenario information available to the agent and $u_{<t}$ the
dialogue history visible before its action at turn $t$. The resulting
\emph{interaction context} at turn $t$ is $h_t=(x,u_{<t})$. Given $h_t$, the
agent selects a social strategy $s_t\in\mathcal{S}(h_t)$, where
$\mathcal{S}(h_t)$ is the set of plausible strategies at that context. The
selected strategy is expressed as a concrete utterance
$u_t\sim\pi(\cdot\mid h_t,s_t)$. The interaction then continues to a terminal
outcome $y$ with score $g(y)$. Thus, $s_t$ denotes the social function of the
choice, whereas $u_t$ denotes its surface realization.

Our goal is to discover empirical candidates for pivotal social strategies:
strategy choices that improve downstream outcomes relative to plausible
alternatives under the same decision context. Conceptually, this target can
be written as
\[
s_t^\star \in \arg\max_{s\in\mathcal{S}(h_t)}
\mathbb{E}\!\left[g(y)\mid h_t,s\right],
\]
where the expectation is taken under a specified continuation policy. This
idealized objective motivates the local evidence construction that follows:
operationally, SoPivot first requires an observed goal-reaching continuation
and then maximizes the sampled mean among eligible candidates
(Section~\ref{sec:comparison}), retaining the matched sibling evidence for
extracting contrast-supported strategy candidates.

\subsection{Sibling Strategy Branching}
\label{sec:branches}

The first stage creates controlled evidence for comparing alternative social
strategies. SoPivot completes a source interaction, selects up to two
candidate decision points, and constructs functionally distinct sibling
strategies under the original visible context. For each sibling strategy,
SoPivot executes the candidate action once at the selected point, samples
repeated natural continuations, and scores the resulting outcomes. This
stage produces scored branch evidence, not yet an agent-facing reusable strategy.

\paragraph{Decision Point Identification.}
Not every turn presents a meaningful strategy choice, so comparing
alternatives at every turn is unnecessary. Let $T=(x,u_{1:L},y)$ denote a
completed source interaction, including its terminal outcome. After the
source interaction is complete, a selector reviews $T$ and returns a set of
up to two target-agent turns, possibly empty:
\[
\mathcal{T}=\operatorname{LLM}_{\mathrm{selector}}(T),\qquad |\mathcal{T}|\le 2.
\]
Each $t^\star\in\mathcal{T}$ is compared independently from its original
prefix; branch continuations are not recursively expanded. The selector is
asked to choose turns where changing the social strategy could substantially
change goal progress, rather than an early greeting or a paraphrase choice.
Seeing the completed trajectory helps allocate the branch budget to the
turns most likely to matter, but the selected turn remains only a
\emph{candidate} critical decision point: all sibling construction for a
selected turn begins from the original pre-action context $h_{t^\star}$,
using only information available before the selected action.

\paragraph{Sibling Strategy Construction.}
At each selected context $h_{t^\star}$, SoPivot keeps the observed action
$u^{(0)}=u_{t^\star}$ and proposes two alternatives, yielding $K=3$ sibling
pairs $(s^{(k)},u^{(k)})$. Here $s^{(k)}$ describes a social function and
$u^{(k)}$ realizes it as an action. A verifier requires plausible,
functionally distinct choices rather than paraphrases; failed construction
produces no set. Retaining the observed action anchors the comparison in
actual source behavior. All siblings share the visible prefix, acting model,
partner, and rollout settings; controller roles and the outcome judge are
specified separately in Appendix~\ref{app:evaluation}.

\paragraph{Local Intervention and Natural Continuation.}
To make branch outcomes comparable, every branch uses the same continuation
policy and rollout settings. For each candidate, SoPivot samples $M$
continuations,
\[
y_{k,m}\sim P_{\mathrm{roll}}(\cdot\mid h_{t^\star},u^{(k)}),
\qquad m=1,\ldots,M.
\]
The action $u^{(k)}$ realizes the candidate strategy at the selected point;
after this action, both agents continue under matched base prompts without
an additional instruction to keep following that strategy. The resulting evidence therefore characterizes this specific action realization under natural, unscripted continuation, rather than a prescribed multi-turn plan. The reported
source bank uses $M=4$ continuations per candidate, and each terminal
outcome is scored as $g(y_{k,m})$ under matched rollout conditions and the
same outcome judge.

\subsection{Contrast-Supported Strategy Identification}
\label{sec:comparison}

Branch outcomes first identify supported candidates; extraction then converts
their evidence into reusable text under the decision-time information boundary.

\paragraph{Branch Outcome Comparison.}
The observed strategy and generated alternatives are treated symmetrically.
For each candidate, SoPivot estimates
\[
\widehat{\mu}_k=\frac{1}{M}\sum_{m=1}^{M}g(y_{k,m})
\]
and retains its full empirical outcome distribution. A success threshold
$\tau=7$ is fixed for source selection: a candidate is eligible if at least
one continuation reaches that threshold, and among eligible candidates,
SoPivot retains all exact ties for the highest unrounded mean:
\[
\begin{aligned}
E&=\{k:\max_m g(y_{k,m})\ge\tau\},\\
W&=\operatorname*{arg\,max}_{k\in E}\widehat{\mu}_k.
\end{aligned}
\]
This two-stage rule prioritizes actions with an observed route to goal
attainment---at least one sampled continuation must reach Goal~$\ge 7$---and
then ranks eligible actions by mean Goal, allowing a sampled successful
route to be retained over consistently subthreshold alternatives. If $E$ is
empty, no strategy is retained, and missing or unjudgeable outcomes for any
sibling make the comparison incomplete and exclude it, though all attempted
continuations remain in the audit record. We call each retained candidate \emph{contrast-supported}: its status rests on the sampled outcome comparison among siblings at this decision point, which is the evidence unit this paper reports throughout. Appendix~\ref{app:success-gate}
examines the success gate and its observed effects.

\paragraph{Evidence-Preserving Strategy Extraction.}
Each winner's evidence is preserved before extraction. One joint extraction
per accepted sibling set uses the complete decision-time visible context,
sibling actions, and downstream process evidence to produce a shared trigger
and a situated strategy for each winner:
\[
\begin{aligned}
\widetilde{s}=(&\text{visible use condition},\text{strategy move},\\
              &\text{expected interaction shift}).
\end{aligned}
\]
The extractor cannot re-rank winners or describe an exact co-winner as better
than another. Process evidence can inform the expected interaction shift,
but future dialogue, source-specific private details, scores, and judge
rationales cannot enter the reusable text, and use conditions and triggers
must depend only on agent-visible information. Invalid extraction yields
no memory entry while preserving the winner evidence.

\subsection{Situated Strategy Management}
\label{sec:memory}

SoPivot retains situated strategies and abstracts compatible entries into
general strategies, without discarding the original situated entries or
their supporting evidence.

\paragraph{Situated and General Strategies.}
The situated memory $\mathcal{M}_{\mathrm{sit}}$ stores each extracted
strategy with its immutable source evidence. A simple aggregation operator
forms a group $G\subseteq\mathcal{M}_{\mathrm{sit}}$ only when the
strategies' visible triggers and strategy moves are functionally compatible,
then removes source-specific entities and reasons to produce a general
strategy $a(G)$. Situated strategies with distinct conditions remain
separate, so the memory does not collapse incompatible cases into one
abstraction.

\paragraph{Abstraction Replay.}
SoPivot evaluates each newly abstracted general strategy against no memory
by restarting whole episodes in its supporting source settings, with the
general strategy optionally available to the focal agent throughout replay.
Acting and partner models, judge, and repeat counts are matched; this
whole-episode replay differs from the one-action discovery intervention
above. Let $\widehat{\mu}^{\mathrm{gen}}$ and $\widehat{\mu}^{0}$ be the mean
Goal scores pooled across all supporting settings. With complete evidence
for both conditions, the admission rule is
\[
\widehat{\mu}^{\mathrm{gen}}\ge\widehat{\mu}^{0}.
\]
Reuse in new scenarios is evaluated in Section~\ref{sec:experiments};
Appendix~\ref{app:expanded-bank} describes the general bank used in the
main experiment.

\paragraph{Memory Update and Retrieval.}
The full memory is
$\mathcal{M}=\mathcal{M}_{\mathrm{sit}}\cup\mathcal{M}_{\mathrm{gen}}$.
Construction and consolidation occur offline; the two banks are frozen before
target evaluation and are not updated from target outcomes. At the start of
a target interaction, one query is constructed from the focal agent's visible
scenario, profile, and goal, and semantic retrieval returns a fixed budget
of multiple situated and general strategies. These are included as
conditional references, not mandatory actions: the focal agent may select,
combine, or ignore them as the dialogue develops, and retrieval is not
repeated at every turn. Source utterances, branch outcomes, and judge
information are never shown to the focal agent.

\section{Experiments}
\label{sec:experiments}

\subsection{Experimental Setup}
\paragraph{Datasets.}
We construct memory from 170 SOTOPIA-\(\pi\) scenarios and use 30 others
for development \citep{wang2024sotopiapi}. The frozen bank contains 272
situated and 34 source-replay-admitted general strategies, and the main
deployment uses up to five situated and three general entries. Evaluation
covers 90 SOTOPIA scenarios with five character pairs each
\citep{zhou2024sotopia}. For each pair, we evaluate the focal agent in each
role separately, yielding $90\times5\times2=900$ configurations per method,
including 140 from SOTOPIA-Hard; only the focal agent's scores enter the
reported results. Appendix~\ref{app:evaluation} gives construction,
deployment, and aggregation details.

\paragraph{Baselines.}
We compare with no memory, dialogue summaries, direct advice,
ReasoningBank, MaTTS, ACE, and PRINCIPLES
\citep{ouyang2026reasoningbank,zhang2025ace,kim-etal-2025-principles}.
Dialogue summaries and direct advice do not use local branches, while MaTTS
contrasts complete source trajectories; all methods share target
configurations and final evaluation, though their construction and online
feedback differ. Appendix~\ref{app:baselines} describes the adaptations and
deployment rules.

\paragraph{Models.}
Qwen3.5-Flash (Flash) serves as target and partner, with temperature 0,
a 500-token action limit, and at most 20 turns. DeepSeek-V4-Pro (DeepSeek)
handles selection, extraction, and applicability checks. Qwen3.8-Max (Max)
is the primary outcome judge, and DeepSeek also rescores final dialogues.
SoPivot retrieves optional memory once for the focal agent; the partner
receives none. Construction settings and prompts appear in
Appendices~\ref{app:evaluation} and~\ref{app:prompts}.

\paragraph{Evaluation.}
Goal (0--10) is the primary metric, and we also report Avg, the arithmetic
mean of the seven raw SOTOPIA dimension scores. Each final dialogue is
scored three times. SoPivot variants share the same generation and scoring
protocol, with up to three dialogue draws per configuration, while external
baselines retain one dialogue per configuration. We average ratings, then
available dialogues within configurations, and weight configurations within
scenes and scenes equally. Appendix~\ref{app:evaluation} details the protocol. Hard is a subset of
All.

\subsection{Main Results on SOTOPIA}
\begin{table*}[!t]
\centering
\sopivottablestyle
\begin{tabular*}{\linewidth}{@{\extracolsep{\fill}}lrrrrrrrr@{}}
\toprule
 & \multicolumn{4}{c}{\textbf{Qwen3.8-Max}} & \multicolumn{4}{c}{\textbf{DeepSeek-V4-Pro}} \\
\cmidrule(lr){2-5}\cmidrule(l){6-9}
 & \multicolumn{2}{c}{\textbf{All (90)}} & \multicolumn{2}{c}{\textbf{Hard (14)}} & \multicolumn{2}{c}{\textbf{All (90)}} & \multicolumn{2}{c}{\textbf{Hard (14)}} \\
\cmidrule(lr){2-3}\cmidrule(lr){4-5}\cmidrule(lr){6-7}\cmidrule(l){8-9}
\textbf{Method} & \textbf{Goal $\uparrow$} & \textbf{Avg $\uparrow$} & \textbf{Goal $\uparrow$} & \textbf{Avg $\uparrow$} & \textbf{Goal $\uparrow$} & \textbf{Avg $\uparrow$} & \textbf{Goal $\uparrow$} & \textbf{Avg $\uparrow$} \\
\midrule
No memory & 6.624 & 2.739 & 5.060 & 1.866 & 6.341 & 2.702 & 4.393 & 1.718 \\
Dialogue summary & 6.595 & 2.772 & 5.102 & 1.968 & 6.280 & 2.719 & 4.367 & 1.793 \\
Direct advice & 6.657 & 2.787 & 4.905 & 1.967 & 6.340 & 2.743 & 4.317 & 1.837 \\
ReasoningBank & 6.704 & 2.914 & 4.993 & 2.234 & 6.358 & 2.861 & 4.271 & 2.023 \\
MaTTS & \underline{6.942} & \underline{3.067} & \textbf{5.452} & \underline{2.357} & 6.448 & \underline{2.927} & \underline{4.652} & \underline{2.103} \\
ACE & 6.883 & 2.975 & 5.174 & 2.233 & \underline{6.498} & 2.874 & 4.495 & 2.013 \\
PRINCIPLES$^{\dagger}$ & 5.727 & 2.253 & 4.243 & 1.881 & 5.982 & 2.574 & 4.324 & 2.052 \\
SoPivot (Ours) & \textbf{7.220} & \textbf{3.117} & \underline{5.399} & \textbf{2.466} & \textbf{6.822} & \textbf{2.995} & \textbf{4.986} & \textbf{2.240} \\
\bottomrule
\end{tabular*}
\caption{Results on SOTOPIA-All and SOTOPIA-Hard under two judges. Goal is the primary metric; Avg averages seven raw dimension scores. Bold/underlining mark the best/second-best results (ties included). $\dagger$: one missing configuration. Evaluation details are in Appendix~\ref{app:evaluation}.}
\label{tab:main-results}
\end{table*}

\paragraph{Learning strategies from local outcomes.}
Under the primary Max judge, SoPivot achieves All Goal 7.220 and Avg 3.117,
compared with 6.624 and 2.739 without memory (Table~\ref{tab:main-results}),
and it also exceeds direct advice and dialogue summaries on Goal on both
All and Hard. These two baselines test simpler alternatives to the proposed
pipeline: producing guidance from the initial source context alone, and
extracting lessons from a completed source experience without local sibling
outcomes. The observed gains support the complete strategy-learning system
over both simpler forms of memory construction.

\paragraph{Comparison with experience-memory methods.}
MaTTS provides a stronger comparison, as it contrasts complete source
trajectories rather than isolated turns. SoPivot improves All Goal over
MaTTS by 0.278 under Max and 0.374 under DeepSeek. Max slightly favors
MaTTS on Hard Goal (5.452 versus 5.399), while SoPivot has higher Avg under
both judges on both subsets. Appendix~\ref{app:baselines} describes how each method is adapted,
including its construction and deployment settings. Section~\ref{sec:analysis}
examines the source evidence and memory components, and
Section~\ref{sec:judge-sensitivity} discusses evaluator sensitivity.

\subsection{Cross-Benchmark Evaluation on AgentSense}
\begin{table}[t]
\centering
\sopivottablestyle
\begin{tabular*}{\linewidth}{@{\extracolsep{\fill}}lrrrr@{}}
\toprule
\textbf{Method} & \textbf{Goal $\uparrow$} & \textbf{Info $\uparrow$} & \textbf{Self $\uparrow$} & \textbf{Other $\uparrow$} \\
\midrule
No memory & 87.050 & 86.100 & 88.950 & 87.650 \\
MaTTS & \underline{87.600} & \underline{87.120} & \textbf{89.970} & \textbf{88.710} \\
SoPivot (Ours) & \textbf{89.000} & \textbf{88.250} & \underline{89.550} & \underline{88.550} \\
\bottomrule
\end{tabular*}
\caption{AgentSense results (\%): 730 two-person scenes, 146 templates and 1,460 focal-role configurations per method. Protocol and metric definitions are in Appendix~\ref{app:agentsense-setup}.}
\label{tab:agentsense}
\end{table}

We evaluate 730 two-person scenes from 146 templates, with both roles
serving as focal agents, yielding 1,460 configurations per method. Both
participants use Flash, with temperature 1 and a 15-message limit. Frozen
source memories transfer without learning from AgentSense outcomes, and
only the focal agent receives retrieved guidance; each nonempty-memory
configuration uses one dialogue, while empty-memory cases reuse matched
no-memory dialogues.

DeepSeek judges external goal completion. Info measures private-information
inference accuracy, while Self and Other report the participants' own
judgments of the focal agent's goal completion. Goal, Self, and Other are
averaged equally across templates, and Info follows the benchmark's
aggregation on available questions (Appendix~\ref{app:agentsense-setup}).

On AgentSense \citep{mou2025agentsense}, SoPivot achieves Goal 89.00\%,
exceeding no memory by 1.95 percentage points and MaTTS by 1.40 points
(Table~\ref{tab:agentsense}). SoPivot also has the highest Info score,
whereas MaTTS has higher Self and Other scores.
Together, these results support cross-benchmark reuse of the learned
memory, with the receiving model fixed to Flash.

\section{Analysis}
\label{sec:analysis}
We organize the analysis around three questions: the role of source outcome
evidence, the contribution of the two memory levels, and sensitivity to the
final evaluation model. Table~\ref{tab:ablation-results} collects the
primary-judge ablations, and Table~\ref{tab:main-results} supplies the
two-judge comparison.

\subsection{Effect of Source Outcome Evidence}
\begin{table}[t]
\centering
\sopivottablestyle
\setlength{\tabcolsep}{3pt}
\begin{tabular*}{\linewidth}{@{\extracolsep{\fill}}lrrrr@{}}
\toprule
 & \multicolumn{2}{c}{\textbf{All}} & \multicolumn{2}{c}{\textbf{Hard}} \\
\cmidrule(lr){2-3}\cmidrule(l){4-5}
\textbf{Variant} & \textbf{Goal $\uparrow$} & \textbf{Avg $\uparrow$} & \textbf{Goal $\uparrow$} & \textbf{Avg $\uparrow$} \\
\midrule
No memory & 6.624 & 2.739 & 5.060 & 1.866 \\
SoPivot (full) & \textbf{7.220} & \textbf{3.117} & \textbf{5.399} & \textbf{2.466} \\
\midrule
\multicolumn{5}{@{}l}{\textit{Memory-level ablations (same bank)}} \\
Situated only & 6.891 & 2.858 & 5.137 & 1.992 \\
General only & 7.131 & 3.064 & 5.300 & 2.410 \\
\midrule
\multicolumn{5}{@{}l}{\textit{Source construction (rebuilt bank)}} \\
w/o source rollouts & 6.937 & 2.995 & 5.210 & 2.220 \\
\bottomrule
\end{tabular*}
\caption{Ablations under Max. Memory-level variants disable one level;
the no-rollout variant rebuilds memory without source continuation outcomes.
Full-system scores match Table~\ref{tab:main-results}.}
\label{tab:ablation-results}
\label{tab:no-source-rollout}
\end{table}

Does observing candidate continuations help build reusable memory beyond
the model's direct judgment? To test this, we retain the source decision
points and three candidate actions but remove source continuations and
their outcomes; DeepSeek instead chooses an action from the decision-time
visible context and extracts a strategy without continuation evidence.

Replacing rollout-based construction lowers Max Goal from 7.220 to 6.937 on
All and from 5.399 to 5.210 on Hard, with Avg decreasing by 0.122 and 0.246,
respectively. This direction is consistent across both subsets and both
metrics, supporting the use of observed downstream outcomes during memory
construction over a direct assessment of plausible actions alone.

Because the variant changes how memory is constructed, its bank is rebuilt under the same pipeline (310 situated and 37 general strategies, versus 272 and 34 in SoPivot), and the comparison covers both bank construction and downstream retrieval. Appendix~\ref{app:no-source-rollout} gives the construction and evaluation details.

\subsection{Complementarity of Memory Levels}
\label{sec:ablation}
Do general strategies replace the need for situated memory? We disable one
level at a time while fixing the bank, scene--role configurations, the
enabled level's retrieval rules, generation, and scoring protocol; unlike
the source-construction comparison above, these variants use the same
frozen strategy bank.

General-only exceeds situated-only on Goal under both judges on All and
Hard: under Max on All, their Goal scores are 7.131 and 6.891, respectively.
Combining the two levels raises Goal to 7.220, adding 0.089 over
general-only and 0.329 over situated-only, and the full system also has
higher Avg than either single-level variant on Hard and under DeepSeek
(Appendix~\ref{app:component-results}).

The two levels keep different things from the sibling comparison that produced them. A situated strategy records the specific visible use condition, strategy move, and expected interaction shift tied to that comparison. A general strategy instead removes source-specific entities, keeping only the trigger and move that compatible situated cases share, so one abstraction can represent a family of similar decision contexts rather than a single scenario. Both remain optional references at retrieval time, tested by enabling one level at a time on a fixed bank.

These results support retaining both specific experiences and their generalizations as optional references: each level contributes complementary guidance, and combining them yields the largest gain. Appendix~\ref{app:source-diagnostic} further compares the two levels at individual source decision points, where both raise Goal over no guidance.

\subsection{Sensitivity to the Evaluation Model}
\label{sec:judge-sensitivity}
The second judge changes evaluation of the collected dialogues rather than
the receiving agent or its memory. Both Max and DeepSeek assign higher Goal
and Avg to SoPivot than to no memory on All and Hard
(Table~\ref{tab:main-results}): the All Goal gains are 0.596 and 0.481,
respectively, and on Hard they are 0.339 and 0.593. SoPivot has the highest
Avg in all four judge--subset settings.

On Hard Goal specifically, Max ranks MaTTS slightly above SoPivot, while DeepSeek ranks SoPivot above MaTTS; both judges nonetheless agree that SoPivot improves over no memory and attains the highest Avg across All and Hard, which is the comparison we treat as the primary signal here.

Dimension-level results are reported in Appendix~\ref{app:full-results}, and Appendix~\ref{app:robustness} gives SoPivot's seven-dimension profile under both judges.

Together with the AgentSense evaluation, these analyses address four questions about SoPivot's design. The source-evidence ablation asks whether rollout-based construction helps, finding it outperforms memory built from a model's direct judgment of the same candidates. The memory-level ablation asks how to combine the two grains for reuse, finding that combining them outperforms either alone. The second judge checks sensitivity to the scoring model, and gains over no memory hold under both Max and DeepSeek. AgentSense asks whether transferred frozen memories raise Goal on a separate benchmark without further learning, and they do.

\section{Conclusion}
We presented SoPivot, which compares the downstream outcomes of candidate
actions under the same interaction context to extract situated and general
strategy memories for selective retrieval and reuse. On SOTOPIA, SoPivot
improves goal achievement over no memory on both All and Hard under two
judges and achieves the highest Avg across all four judge--subset settings.
Comparisons with direct advice, dialogue summaries, and trajectory-level
memory methods support the effectiveness of this strategy-learning pipeline,
while component experiments show the complementary value of situated and
general strategies. Results on AgentSense further demonstrate the potential
for cross-benchmark reuse. Together, these findings suggest that distilling local interaction outcomes into reusable strategies is an effective approach to improving language models' social interaction performance, complementing trajectory-level and advice-based memory with evidence grounded in local, context-matched comparisons.

\clearpage

\section*{Limitations}
This study evaluates SoPivot through simulated social interactions and
model-based scoring; human evaluation and a broader range of interaction
environments would help establish its applicability in real-world settings. Strategy discovery relies on a finite set of candidate
actions and sampled continuations, so its effectiveness may vary across
models and interaction conditions. Methods also differ in the cost of
memory construction and use, and comparing their performance and
efficiency under a shared computational budget remains an open question.

\section*{Ethical Considerations}
This study uses simulated agents and does not recruit new human
participants. It investigates memory-guided social interaction; results
from simulated evaluations do not establish suitability for direct
interaction with people. Reusable social strategies carry a risk of misuse
for manipulation, and applications involving real users would require
further assessment of privacy, informed consent, and the interests of both
interaction partners. Memory and retrieval in this work respect the
agent-visible information boundary.

\paragraph{AI assistance.}
Generative AI tools supported research planning, literature search, code
development, and manuscript preparation. The authors are responsible for
verifying all methods, experimental results, citations, and text.

 \bibliography{references}

\begin{thebibliography}{30}
\providecommand{\natexlab}[1]{#1}

\bibitem[{Chao et~al.(2026)Chao, Bai, Sheng, Li, and Sun}]{chao2026stale}
Hanxiang Chao, Yihan Bai, Rui Sheng, Tianle Li, and Yushi Sun. 2026.
\newblock \href {https://arxiv.org/abs/2605.06527} {{STALE}: Can {LLM} agents
  know when their memories are no longer valid?}
\newblock \emph{Preprint}, arXiv:2605.06527.

\bibitem[{Chawla et~al.(2021)Chawla, Ramirez, Clever, Lucas, May, and
  Gratch}]{chawla2021casino}
Kushal Chawla, Jaysa Ramirez, Rene Clever, Gale Lucas, Jonathan May, and
  Jonathan Gratch. 2021.
\newblock \href {https://doi.org/10.18653/v1/2021.naacl-main.254} {{CaSiNo}: A
  corpus of campsite negotiation dialogues for automatic negotiation systems}.
\newblock In \emph{Proceedings of the 2021 Conference of the North American
  Chapter of the Association for Computational Linguistics: Human Language
  Technologies}, pages 3167--3185. Association for Computational Linguistics.

\bibitem[{Chen et~al.(2026)Chen, Zhong, Liu, and Du}]{chen2026skillcat}
Kunfeng Chen, Qihuang Zhong, Juhua Liu, and Bo~Du. 2026.
\newblock \href {https://arxiv.org/abs/2606.13317v1} {{SkillCAT}: Contrastive
  assessment and topology-aware skill self-evolution for {LLM} agents}.
\newblock \emph{Preprint}, arXiv:2606.13317v1.

\bibitem[{Gao et~al.(2026)Gao, Chen, Wang, Guo, Pang, Liu, Shen, and
  Cheng}]{gao2026skillaudit}
Haowen Gao, Haoran Chen, Can Wang, Shasha Guo, Liang Pang, Zhaoyang Liu, Huawei
  Shen, and Xueqi Cheng. 2026.
\newblock \href {https://arxiv.org/abs/2606.14239v1} {{SkillAudit}:
  Ground-truth-free skill evolution via paired trajectory auditing}.
\newblock \emph{Preprint}, arXiv:2606.14239v1.

\bibitem[{Goel and Zhu(2025)}]{goel2025lifelongsotopia}
Hitesh Goel and Hao Zhu. 2025.
\newblock \href {https://arxiv.org/abs/2506.12666} {{LIFELONG SOTOPIA}:
  Evaluating social intelligence of language agents over lifelong social
  interactions}.
\newblock \emph{Preprint}, arXiv:2506.12666.

\bibitem[{Kim et~al.(2025)Kim, Ong, Hwang, Kang, Jihn, Kim, Kim, and
  Yeo}]{kim-etal-2025-principles}
Namyoung Kim, Kai Tzu-iunn Ong, Yeonjun Hwang, Minseok Kang, Iiseo Jihn,
  Gayoung Kim, Minju Kim, and Jinyoung Yeo. 2025.
\newblock \href {https://doi.org/10.18653/v1/2025.findings-emnlp.1164}
  {{PRINCIPLES}: Synthetic strategy memory for proactive dialogue agents}.
\newblock In \emph{Findings of the Association for Computational Linguistics:
  EMNLP 2025}, pages 21329--21368. Association for Computational Linguistics.

\bibitem[{{Kimi Team} et~al.(2026){Kimi Team}, Bai, Bai, Bao, {M. C.}, Cai,
  Cai, Cao, Cao, Chai, {Y. Charles} et~al.}]{team2026kimi}
{Kimi Team}, Tongtong Bai, Yifan Bai, Yiping Bao, {M. C.}, Jianfeng Cai,
  Xinyuan Cai, Peizhou Cao, Yuxuan Cao, Ziwei Chai, {Y. Charles}, et~al. 2026.
\newblock \href {https://arxiv.org/abs/2607.24653} {{Kimi K3}: Open frontier
  intelligence}.
\newblock \emph{Preprint}, arXiv:2607.24653.

\bibitem[{Lei et~al.(2026)Lei, Wang, Lian, Hu, Lian, and Xie}]{lei2026humanllm}
Yuxuan Lei, Tianfu Wang, Jianxun Lian, Zhengyu Hu, Defu Lian, and Xing Xie.
  2026.
\newblock \href {https://doi.org/10.1145/3770854.3780294} {{HumanLLM}: Towards
  personalized understanding and simulation of human nature}.
\newblock In \emph{Proceedings of the 32nd ACM SIGKDD Conference on Knowledge
  Discovery and Data Mining V.1}, pages 621--632. Association for Computing
  Machinery.

\bibitem[{Lewis et~al.(2017)Lewis, Yarats, Dauphin, Parikh, and
  Batra}]{lewis2017deal}
Mike Lewis, Denis Yarats, Yann Dauphin, Devi Parikh, and Dhruv Batra. 2017.
\newblock \href {https://doi.org/10.18653/v1/D17-1259} {Deal or no deal?
  end-to-end learning of negotiation dialogues}.
\newblock In \emph{Proceedings of the 2017 Conference on Empirical Methods in
  Natural Language Processing}, pages 2443--2453. Association for Computational
  Linguistics.

\bibitem[{Liu et~al.(2021)Liu, Zheng, Demasi, Sabour, Li, Yu, Jiang, and
  Huang}]{liu2021esc}
Siyang Liu, Chujie Zheng, Orianna Demasi, Sahand Sabour, Yu~Li, Zhou Yu, Yong
  Jiang, and Minlie Huang. 2021.
\newblock \href {https://doi.org/10.18653/v1/2021.acl-long.269} {Towards
  emotional support dialog systems}.
\newblock In \emph{Proceedings of the 59th Annual Meeting of the Association
  for Computational Linguistics and the 11th International Joint Conference on
  Natural Language Processing (Volume 1: Long Papers)}, pages 3469--3483.
  Association for Computational Linguistics.

\bibitem[{Mou et~al.(2025)Mou, Liang, Lin, Zhang, Liu, Yang, Ye, Chen, Kuang,
  Huang, and Wei}]{mou2025agentsense}
Xinyi Mou, Jingcong Liang, Jiayu Lin, Xinnong Zhang, Xiawei Liu, Shiyue Yang,
  Rong Ye, Lei Chen, Haoyu Kuang, Xuanjing Huang, and Zhongyu Wei. 2025.
\newblock \href {https://doi.org/10.18653/v1/2025.naacl-long.257}
  {{AgentSense}: Benchmarking social intelligence of language agents through
  interactive scenarios}.
\newblock In \emph{Proceedings of the 2025 Conference of the Nations of the
  Americas Chapter of the Association for Computational Linguistics: Human
  Language Technologies (Volume 1: Long Papers)}, pages 4975--5001. Association
  for Computational Linguistics.

\bibitem[{Ouyang et~al.(2026)Ouyang, Yan, Hsu, Chen, Jiang, Wang, Han, Le,
  Daruki, Tang, Tirumalashetty, Lee, Rofouei, Lin, Han, Lee, and
  Pfister}]{ouyang2026reasoningbank}
Siru Ouyang, Jun Yan, I-Hung Hsu, Yanfei Chen, Ke~Jiang, Zifeng Wang, Rujun
  Han, Long~T. Le, Samira Daruki, Xiangru Tang, Vishy Tirumalashetty, George
  Lee, Mahsan Rofouei, Hangfei Lin, Jiawei Han, Chen-Yu Lee, and Tomas Pfister.
  2026.
\newblock \href {https://openreview.net/forum?id=jL7fwchScm} {{ReasoningBank}:
  Scaling agent self-evolving with reasoning memory}.
\newblock In \emph{The Fourteenth International Conference on Learning
  Representations}.

\bibitem[{Park et~al.(2023)Park, O'Brien, Cai, Morris, Liang, and
  Bernstein}]{park2023generative}
Joon~Sung Park, Joseph O'Brien, Carrie~J. Cai, Meredith~Ringel Morris, Percy
  Liang, and Michael~S. Bernstein. 2023.
\newblock \href {https://doi.org/10.1145/3586183.3606763} {Generative agents:
  Interactive simulacra of human behavior}.
\newblock In \emph{Proceedings of the 36th Annual ACM Symposium on User
  Interface Software and Technology}, pages 1--22. Association for Computing
  Machinery.

\bibitem[{Pu et~al.(2026)Pu, Song, and Zhao}]{pu2026skillops}
Hongji Pu, Xinyuan Song, and Liang Zhao. 2026.
\newblock \href {https://arxiv.org/abs/2605.13716} {{SkillOps}: Managing {LLM}
  agent skill libraries as self-maintaining software ecosystems}.
\newblock \emph{Preprint}, arXiv:2605.13716.

\bibitem[{Shi et~al.(2026)Shi, Chen, Li, Pan, Sun, Xu, Zhou, and
  Li}]{shi2026r3l}
Weijie Shi, Yanxi Chen, Zexi Li, Xuchen Pan, Yuchang Sun, Jiajie Xu, Xiaofang
  Zhou, and Yaliang Li. 2026.
\newblock \href {https://arxiv.org/abs/2601.03715} {{$R^3L$}:
  Reflect-then-retry reinforcement learning with language-guided exploration,
  pivotal credit, and positive amplification}.
\newblock \emph{Preprint}, arXiv:2601.03715.

\bibitem[{Shinn et~al.(2023)Shinn, Cassano, Gopinath, Narasimhan, and
  Yao}]{shinn2023reflexion}
Noah Shinn, Federico Cassano, Ashwin Gopinath, Karthik Narasimhan, and Shunyu
  Yao. 2023.
\newblock \href {https://doi.org/10.52202/075280-0377} {Reflexion: Language
  agents with verbal reinforcement learning}.
\newblock In \emph{Advances in Neural Information Processing Systems},
  volume~36, pages 8634--8652.

\bibitem[{Wang et~al.(2024{\natexlab{a}})Wang, Xie, Jiang, Mandlekar, Xiao,
  Zhu, Fan, and Anandkumar}]{wang2024voyager}
Guanzhi Wang, Yuqi Xie, Yunfan Jiang, Ajay Mandlekar, Chaowei Xiao, Yuke Zhu,
  Linxi Fan, and Anima Anandkumar. 2024{\natexlab{a}}.
\newblock \href {https://openreview.net/forum?id=ehfRiF0R3a} {Voyager: An
  open-ended embodied agent with large language models}.
\newblock \emph{Transactions on Machine Learning Research}.

\bibitem[{Wang et~al.(2024{\natexlab{b}})Wang, Yu, Zhang, Qi, Sap, Bisk,
  Neubig, and Zhu}]{wang2024sotopiapi}
Ruiyi Wang, Haofei Yu, Wenxin Zhang, Zhengyang Qi, Maarten Sap, Yonatan Bisk,
  Graham Neubig, and Hao Zhu. 2024{\natexlab{b}}.
\newblock \href {https://doi.org/10.18653/v1/2024.acl-long.698}
  {{SOTOPIA}-{$\pi$}: Interactive learning of socially intelligent language
  agents}.
\newblock In \emph{Proceedings of the 62nd Annual Meeting of the Association
  for Computational Linguistics (Volume 1: Long Papers)}, pages 12912--12940.
  Association for Computational Linguistics.

\bibitem[{Wang et~al.(2026)Wang, Xiong, Lian, Zhu, Hu, Lei, Gong, Hu, Li, Tsai,
  Yuan, and Zhang}]{wang2026socialcoach}
Tianfu Wang, Max Xiong, Jianxun Lian, Hongyuan Zhu, Zhengyu Hu, Yuxuan Lei,
  Linxiao Gong, Dapeng Hu, Xiaofang Li, Peiting Tsai, Nicholas~Jing Yuan, and
  Qi~Zhang. 2026.
\newblock \href {https://arxiv.org/abs/2606.04155} {{SocialCoach}: Personalized
  social skill learning with agentic tutoring and practice}.
\newblock \emph{Preprint}, arXiv:2606.04155.

\bibitem[{Wang et~al.(2025{\natexlab{a}})Wang, Zhan, Lian, Hu, Yuan, Zhang,
  Xie, and Xiong}]{wang2025llm}
Tianfu Wang, Yi~Zhan, Jianxun Lian, Zhengyu Hu, Nicholas~Jing Yuan, Qi~Zhang,
  Xing Xie, and Hui Xiong. 2025{\natexlab{a}}.
\newblock \href {https://doi.org/10.1145/3701716.3715244} {{LLM}-powered
  multi-agent framework for goal-oriented learning in intelligent tutoring
  system}.
\newblock In \emph{Companion Proceedings of the ACM on Web Conference 2025},
  pages 510--519. Association for Computing Machinery.

\bibitem[{Wang et~al.(2025{\natexlab{b}})Wang, Mao, Fried, and
  Neubig}]{wang2025awm}
Zora~Zhiruo Wang, Jiayuan Mao, Daniel Fried, and Graham Neubig.
  2025{\natexlab{b}}.
\newblock \href {https://proceedings.mlr.press/v267/wang25bx.html} {Agent
  workflow memory}.
\newblock In \emph{Proceedings of the 42nd International Conference on Machine
  Learning}, volume 267 of \emph{Proceedings of Machine Learning Research},
  pages 63897--63911. PMLR.

\bibitem[{Yang et~al.(2026{\natexlab{a}})Yang, Liu, Huang, Wu, Lei, and
  Ng}]{yang2026metro}
Haofu Yang, Jiaji Liu, Chen Huang, Faguo Wu, Wenqiang Lei, and See-Kiong Ng.
  2026{\natexlab{a}}.
\newblock \href {https://arxiv.org/abs/2604.11427} {{METRO}: Towards strategy
  induction from expert dialogue transcripts for non-collaborative dialogues}.
\newblock \emph{Preprint}, arXiv:2604.11427.

\bibitem[{Yang et~al.(2026{\natexlab{b}})Yang, Gong, Huang, Yang, Zhou, Huang,
  Li, Gao, Dai, Liu, Qiu, Yang, Chen, Yang, and Luo}]{yang2026skillopt}
Yifan Yang, Ziyang Gong, Weiquan Huang, Qihao Yang, Ziwei Zhou, Zisu Huang, Yan
  Li, Xuemei Gao, Qi~Dai, Bei Liu, Kai Qiu, Yuqing Yang, Dongdong Chen, Xue
  Yang, and Chong Luo. 2026{\natexlab{b}}.
\newblock \href {https://arxiv.org/abs/2605.23904} {{SkillOpt}: Executive
  strategy for self-evolving agent skills}.
\newblock \emph{Preprint}, arXiv:2605.23904.

\bibitem[{Yu et~al.(2025)Yu, Qi, Zhao, Nottingham, Xuan, Majumder, Zhu, Liang,
  and You}]{yu2025sotopiarl}
Haofei Yu, Zhengyang Qi, Yining Zhao, Kolby Nottingham, Keyang Xuan,
  Bodhisattwa~Prasad Majumder, Hao Zhu, Paul~Pu Liang, and Jiaxuan You. 2025.
\newblock \href {https://arxiv.org/abs/2508.03905} {{Sotopia-RL}: Reward design
  for social intelligence}.
\newblock \emph{Preprint}, arXiv:2508.03905.

\bibitem[{Zhang et~al.(2026{\natexlab{a}})Zhang, Sun, Su, Yang, Pan, Chen, and
  Zhang}]{zhang2026mica}
Naifan Zhang, Ruihan Sun, Jinwei Su, Hengjie Yang, Zhengyuan Pan, Zhaohan Chen,
  and Xiaofan Zhang. 2026{\natexlab{a}}.
\newblock \href {https://arxiv.org/abs/2603.06194} {{MICA}: Multi-granularity
  intertemporal credit assignment for long-horizon emotional support dialogue}.
\newblock \emph{Preprint}, arXiv:2603.06194.

\bibitem[{Zhang et~al.(2026{\natexlab{b}})Zhang, Hu, Upasani, Ma, Hong,
  Kamanuru, Rainton, Wu, Ji, Li, Thakker, Zou, and Olukotun}]{zhang2025ace}
Qizheng Zhang, Changran Hu, Shubhangi Upasani, Boyuan Ma, Fenglu Hong,
  Vamsidhar Kamanuru, Jay Rainton, Chen Wu, Mengmeng Ji, Hanchen Li, Urmish
  Thakker, James Zou, and Kunle Olukotun. 2026{\natexlab{b}}.
\newblock \href {https://doi.org/10.48550/arXiv.2510.04618} {Agentic context
  engineering: Evolving contexts for self-improving language models}.
\newblock In \emph{The Fourteenth International Conference on Learning
  Representations}.

\bibitem[{Zhang et~al.(2025)Zhang, Liu, Song, Li, and
  Liu}]{zhang2025sotopiaomega}
Wenyuan Zhang, Tianyun Liu, Mengxiao Song, Xiaodong Li, and Tingwen Liu. 2025.
\newblock \href {https://doi.org/10.18653/v1/2025.acl-long.1203}
  {{SOTOPIA}-{$\Omega$}: Dynamic strategy injection learning and social
  instruction following evaluation for social agents}.
\newblock In \emph{Proceedings of the 63rd Annual Meeting of the Association
  for Computational Linguistics (Volume 1: Long Papers)}, pages 24669--24697.
  Association for Computational Linguistics.

\bibitem[{Zhao et~al.(2024)Zhao, Huang, Xu, Lin, Liu, and
  Huang}]{zhao2024expel}
Andrew Zhao, Daniel Huang, Quentin Xu, Matthieu Lin, Yong-Jin Liu, and Gao
  Huang. 2024.
\newblock \href {https://doi.org/10.1609/aaai.v38i17.29936} {{ExpeL}: {LLM}
  agents are experiential learners}.
\newblock In \emph{Proceedings of the AAAI Conference on Artificial
  Intelligence}, volume~38, pages 19632--19642.

\bibitem[{Zhou et~al.(2024{\natexlab{a}})Zhou, Yan, Shlapentokh-Rothman, Wang,
  and Wang}]{zhou2023lats}
Andy Zhou, Kai Yan, Michal Shlapentokh-Rothman, Haohan Wang, and Yu-Xiong Wang.
  2024{\natexlab{a}}.
\newblock \href {https://proceedings.mlr.press/v235/zhou24r.html} {Language
  agent tree search unifies reasoning, acting, and planning in language
  models}.
\newblock In \emph{Proceedings of the 41st International Conference on Machine
  Learning}, volume 235 of \emph{Proceedings of Machine Learning Research},
  pages 62138--62160. PMLR.

\bibitem[{Zhou et~al.(2024{\natexlab{b}})Zhou, Zhu, Mathur, Zhang, Qi, Yu,
  Morency, Bisk, Fried, Neubig, and Sap}]{zhou2024sotopia}
Xuhui Zhou, Hao Zhu, Leena Mathur, Ruohong Zhang, Zhengyang Qi, Haofei Yu,
  Louis-Philippe Morency, Yonatan Bisk, Daniel Fried, Graham Neubig, and
  Maarten Sap. 2024{\natexlab{b}}.
\newblock \href {https://openreview.net/forum?id=mM7VurbA4r} {{SOTOPIA}:
  Interactive evaluation for social intelligence in language agents}.
\newblock In \emph{The Twelfth International Conference on Learning
  Representations}.

\end{thebibliography}

\clearpage
\appendix
\raggedbottom
\makeatletter
\setlength{\@fptop}{0pt}
\setlength{\@dblfptop}{0pt}
\makeatother

\section{Additional Method Details}
\label{app:method-details}

This appendix expands the local comparison, joint extraction, and frozen reuse introduced in Section~\ref{sec:method}; evaluation settings are specified in Appendix~\ref{app:evaluation}. Prompts and implementation notes refer to the focal agent as the \emph{student}.

\begingroup
\setlength{\intextsep}{10pt}
\algrenewcommand\algorithmicindent{1em}
\subsection{Algorithms}
\label{app:algorithms}
The pseudocode separates evidence collection, memory admission, and target
reuse. Algorithms~\ref{alg:local-contrast}--\ref{alg:general-admission}
describe memory construction, and Algorithm~\ref{alg:frozen-reuse}
describes deployment with the frozen bank
(Appendix~\ref{app:expanded-bank}). Model calls use the templates in
Appendix~\ref{app:prompts}.

\begin{algorithm}[H]
\caption{Local contrast and situated memory}
\label{alg:local-contrast}
\small
\begin{algorithmic}[1]
\Require Source episodes; $K=3$; $M=4$; $\tau=7$
\Ensure Situated store $S$; evidence audit $A$
\State Initialize $S$ and $A$ as empty stores
\For{each completed source episode $x$}
  \State Select at most two original nodes in $x$
  \For{each selected node $t$}
    \State $h\gets$ student-visible context before $t$
    \State $a_0\gets$ observed action at $t$
    \State Propose two alternatives given $(h,a_0)$
    \State $C\gets\{a_0\}\cup\{\text{two proposed actions}\}$
    \If{$|C|\ne K$ or candidates are not distinct}
      \State \textbf{continue}
    \EndIf
    \For{each $a\in C$ and repeat $r=1,\ldots,M$}
      \State $y_{a,r}\gets$ continue from original $t$ with $a$
      \State $g_{a,r}\gets$ judge Goal score of $y_{a,r}$
    \EndFor
    \State Add $(h,C,y,g)$ to audit $A$
    \If{any required outcome is missing}
      \State \textbf{continue}
    \EndIf
    \State $E\gets\{a\in C:\max_r g_{a,r}\ge\tau\}$
    \If{$E=\varnothing$} \State \textbf{continue} \EndIf
    \State $\mu_a\gets\frac{1}{M}\sum_r g_{a,r}$ for each $a\in E$
    \State $W\gets\arg\max_{a\in E}\mu_a$ (all exact ties)
    \State Jointly extract $d$ from $(h,C,y,W)$
    \State Review $d$ using $(h,C,W)$
    \State Allow at most two repairs
    \If{$d$ passes schema and boundary checks}
      \State Add $d$ with every winner in $W$ to $S$
    \EndIf
  \EndFor
\EndFor
\State \Return $S,A$
\end{algorithmic}
\end{algorithm}
Each continuation restores the original environment state and each actor's
observation. Only the first action is replaced; subsequent actions follow
the matched base policies. The mean includes all four continuations. Exact ties use
unrounded means, and extraction cannot change the selected winner set.
Context review is isolated from candidate actions and future dialogue;
action review receives the selected actions. Invalid extraction leaves
the comparison in the audit but adds no deployable entry.

\begin{algorithm}[H]
\caption{General-strategy admission}
\label{alg:general-admission}
\small
\begin{algorithmic}[1]
\Require Situated store $S$; source settings; $R=3$
\Ensure Frozen situated and general stores $S,G$
\State Recall semantically similar pairs from $S$
\State Initialize the approved-pair set as empty
\For{each recalled pair $(s_1,s_2)$}
  \If{both dilemma and move are compatible}
    \State Approve $(s_1,s_2)$
  \EndIf
\EndFor
\State Initialize $G$ as an empty store
\For{each group formed from approved pairs}
  \State $q\gets$ abstract a general strategy from the group
  \If{$q$ fails schema or boundary checks}
    \State \textbf{continue}
  \EndIf
  \State $I\gets$ distinct supporting source settings of $q$
  \For{each $i\in I$ and repeat $r=1,\ldots,R$}
    \State Restart episode $i$ with general memory $q$
    \State $g^G_{i,r}\gets$ judge Goal score of that episode
    \State Restart matched episode $i$ with no memory
    \State $g^0_{i,r}\gets$ judge Goal score of that episode
  \EndFor
  \If{any matched result is missing}
    \State \textbf{continue}
  \EndIf
  \If{$\operatorname{mean}_{i,r}g^G_{i,r}\ge
       \operatorname{mean}_{i,r}g^0_{i,r}$}
    \State Add $q$ and supporting strategy IDs to $G$
  \EndIf
\EndFor
\State \Return frozen $S,G$
\end{algorithmic}
\end{algorithm}
Admission averages over distinct supporting settings and matched repeats.
It compares general memory with no memory over complete episodes, and
does not require general memory to match situated memory at every node.

\begin{algorithm}[H]
\caption{Frozen episode-start reuse}
\label{alg:frozen-reuse}
\small
\begin{algorithmic}[1]
\Require Initial student-visible context $v$; frozen $S,G$
\Ensure Target interaction
\State $q\gets$ embed the visible query from $v$
\State $C_S\gets$ top 10 semantic matches in $S$
\State $C_G\gets$ all entries in $G$, ordered by similarity
\State $V_S\gets$ entries in $C_S$ applicable to $v$
\State $V_G\gets$ entries in $C_G$ applicable to $v$
\State $B\gets$ up to five entries from $V_S$ and three from
\Statex \hspace{\algorithmicindent}$V_G$, selected by similarity within each store
\If{$\operatorname{MemoryTokens}(B)>12000$}
  \State Stop and report an oversized bundle
\EndIf
\If{$B=\varnothing$}
  \State \Return interaction with the base policy
\EndIf
\State Render $B$ as optional strategy references
\State \Return interaction with the base policy, persistent
\Statex \hspace{\algorithmicindent}student memory $B$, and no partner memory
\end{algorithmic}
\end{algorithm}
The query and applicability check use the initial student view. The agent
checks a strategy's conditions against the actual dialogue before using it;
it may combine or ignore references. No per-turn retrieval or target-outcome
memory update occurs.

\endgroup

\subsection{Local Comparison and Joint Extraction}

The observed action anchors the sibling comparison to behavior that actually
occurred. SoPivot selects up to two points after
reviewing a completed source interaction; selection may abstain. It does not recursively
expand later decisions. The observed action and two alternatives form $K=3$
branches. Each local action is followed by natural continuation under matched
base prompts, without a persistent instruction to follow its strategy.
The comparison therefore concerns concrete action realizations under a
specified continuation policy.

Algorithm 1 in Appendix~\ref{app:algorithms} specifies the construction procedure. Incomplete sibling evidence is rejected rather than assigned a zero score, and all exact co-winners are preserved. Selection uses Goal, while all seven dimensions remain available for audit.

\paragraph{Executed construction budget.}
Across 170 source dialogues, 165 contribute two selected positions and five
contribute one, yielding 335 local comparisons. Each contains three actions
and four continuations per action: 1,005 candidate actions and 4,020 scored
continuations.
Table~\ref{tab:construction-funnel} separates comparison-level decisions from
candidate actions and exported memory entries. A joint extraction can yield
multiple strategies, so these units cannot be subtracted interchangeably.

\begin{table}[tbp]
\centering
\sopivottablestyle
\begin{tabular}{@{}p{0.79\columnwidth}r@{}}
\toprule
\textbf{Construction stage (counting unit)} & \textbf{Count} \\
\midrule
Source dialogues & 170 \\
Completed local comparisons & 335 \\
Comparisons with no eligible action & 26 \\
Comparisons passed to joint extraction & 309 \\
\addlinespace[3pt]
Candidate actions evaluated & 1,005 \\
Actions meeting the success gate & 857 \\
Selected actions, including exact ties & 468 \\
\addlinespace[3pt]
Joint extractions rejected by content review & 131 \\
Other exclusions & 1 \\
Admitted joint extractions & 177 \\
Exported situated strategies & 272 \\
Source scenarios represented in the store & 129 \\
\addlinespace[3pt]
General strategies admitted by source replay & 34 \\
\bottomrule
\end{tabular}
\caption{Construction accounting for the situated and general memory. Of the
309 extraction tasks, 177 are admitted, 131 are rejected by content review,
and one is excluded for other reasons. Appendix~\ref{app:expanded-bank}
describes general-strategy admission.}
\label{tab:construction-funnel}
\end{table}

\subsection{Success-Gated Selection: Rationale and Cases}
\label{app:success-gate}

The success gate implements a two-stage discovery objective: first retain
actions with an observed continuation reaching Goal~$\ge 7$, then select
those with the highest mean among eligible actions. It preserves evidence
of goal attainment that a comparison of means alone may discard. The gate
therefore complements the mean criterion rather than serving as an
unconstrained expected-Goal maximizer.

\paragraph{Selection audit.}
Across 335 completed comparisons, the gate selects a lower-mean action over
an ineligible alternative in three cases (0.90\%). Among the 309 comparisons
with eligible candidates, the selected mean exceeds the observed action's
mean in 157, matches it in 151, and falls below it in one. The latter case
(scenario 138) is excluded by extraction content review. Consequently, none
of the final 272 situated entries comes from a selected action whose mean
is below that of the observed action. Two gate exceptions enter the bank;
both outperform the observed action while preserving an observed successful
path unavailable among the sampled continuations of another alternative.

\paragraph{Illustrative rationale.}
In scenario 012, an agent seeks a roommate's commitment to daily plant
watering. A soil-check compromise yields partial cooperation without the
requested schedule. A friendship-and-trust appeal has a successful
continuation in which the roommate explicitly agrees to daily watering;
other continuations resist the appeal. The gate retains this observed route
to the target outcome for strategy extraction. This illustrates the intended
discovery criterion; downstream memory evaluation separately tests reuse.

\subsection{Evidence and Student-Visible Information}
\label{app:information-boundary}

A retrospective selector may inspect a completed source dialogue to allocate
exploration. Branch comparison uses scores to choose supported candidates.
Neither permission makes future events valid applicability conditions for
reusable memory. The extractor receives the complete decision-time visible
history, sibling actions, downstream process evidence, and already selected
winner IDs. Process evidence can inform an abstract expected interaction
shift, such as making constraints explicit. It cannot authorize copying
future dialogue, hidden partner goals, terminal outcomes, scores, or judge
reasoning into reusable text. The extractor cannot re-rank winners.
Table~\ref{tab:appendix-boundary} summarizes these permissions.

\begin{table}[t]
\centering
\sopivottablestyle
\begin{tabular}{@{}>{\raggedright\arraybackslash}p{0.34\columnwidth}
>{\raggedright\arraybackslash}p{0.61\columnwidth}@{}}
\toprule
\textbf{Information} & \textbf{Permitted use} \\
\midrule
Visible context & Use conditions, trigger, retrieval query, and target action. \\
Sibling actions & Identify the supported strategy move during extraction. \\
Downstream process & Infer an abstract expected shift; no copied future dialogue. \\
Scores and judge output & Selection and audit; excluded from reusable text. \\
Partner-private facts & Excluded from student-facing memory and applicability decisions. \\
\bottomrule
\end{tabular}
\caption{Information boundary. Construction evidence is not automatically
available to the acting student.}
\label{tab:appendix-boundary}
\end{table}

Each situated strategy states a visible use condition, a strategy action,
and an expected interaction shift. Source and process references support
traceability, not extra target-visible facts. Exact co-winners may recommend
distinct moves, but extraction cannot describe one as better than another.

\subsection{Consolidation and General-Strategy Admission}
\label{app:general-admission}

Embedding similarity between situated use conditions and strategy actions
proposes cross-case matches. Similarity alone is insufficient: an adjudicator
must confirm both the same visible interaction dilemma and the same strategy
move. Uncertain or incompatible cases remain separate. Compatible cases can
be abstracted by removing source-specific entities while preserving
supporting strategy IDs and evidence links.

The abstraction receives its own whole-episode source-setting replay check.
Episodes restart from their initial context, with the general strategy
available as optional memory throughout. This is not the local one-action
intervention. Let $I$ be
its distinct supporting settings and $R$ the declared matched repeat count
per condition and setting. The two conditions are the general strategy and
no memory, not the original situated strategy. For $a\in\{\mathrm{gen},0\}$,
\[
\widehat{\mu}^{a}
=\frac{1}{|I|R}\sum_{i\in I}\sum_{r=1}^{R}g(y^{a}_{i,r}).
\]
Complete matched evidence is required, and admission requires
$\widehat{\mu}^{\mathrm{gen}}\ge\widehat{\mu}^{0}$. Missing settings or
conditions constitute insufficient evidence. Situated memory
is supported by local contrast; the source-setting gate applies to newly
abstracted general strategies.

\subsection{Frozen Two-Level Retrieval}

Situated and general strategies remain in separate frozen stores;
new target outcomes do not update them. At episode start, a query uses only
the student's visible scenario, own profile and goal, and available public
information. Semantic retrieval and applicability checking select a bounded
bundle. Conditional references remain available throughout the dialogue;
retrieval is not repeated every turn.

Applicability means that a strategy is worth carrying into the episode.
A dialogue-dependent condition need not already hold, but there must be a
concrete visible task, role, or goal link. Unknown future facts are not
assumed true. Known incompatibility, confidentiality, permissions, and
infeasibility cannot be bypassed by suggesting a condition might arise later.
The student may use, combine, or ignore references as actual conditions
develop; it must not manufacture a situation to fit a strategy.

If nothing qualifies, no strategy is injected. A short bundle is not filled
with substitutes to meet a quota. This interface supports multiple optional
strategies, rather than a single-strategy limit. The executed caps are
listed in Appendix~\ref{app:evaluation}.

\section{Evaluation Protocol}
\label{app:evaluation}

\subsection{Data Roles}
The source pool contains 170 scenarios from a 200-scenario
SOTOPIA-\(\pi\) snapshot; the other 30 serve development. Source dialogues
are generated with the native scenario and character profiles. Source and
development scenario IDs are disjoint from the official 90 evaluation
scenarios. The memory bank and deployment configuration are frozen before
official outcomes are collected.

Construction allows up to two decision points per source dialogue, with
three candidates and four continuations per candidate, and yields 272
situated strategies.

\subsection{General Bank Used in the Main Experiment}
\label{app:expanded-bank}
The main experiment deploys 34 general strategies that pass source-scenario
replay admission, alongside the 272 situated strategies.

Each replay candidate is compared against no memory in its supporting source
settings, with three Flash dialogue draws per arm (temperature 0.7, seeds
1601--1603, 500 output tokens and at most 20 turns). Each dialogue receives
one Max judgment at temperature 0. Admission uses the pooled mean Goal rule
of Appendix~\ref{app:general-admission}.

At episode start, situated retrieval checks its top ten semantic matches and
keeps at most five applicable entries. General retrieval checks all 34
admitted entries and keeps at most three, ordered by similarity. The optional
memory is fixed throughout each episode and visible only to the focal agent.
A 12,000-token guard bounds the memory bundle; the largest bundle has 2,069
estimated tokens. Nonempty memory reaches 751/900 configurations (134/140
Hard), including general memory in 665/900 (129/140 Hard).

Target dialogues use temperature 0 with seeds 1701--1703, with up to three
draws per configuration; configurations whose retrieved bundle is empty use
the matched no-memory dialogue. The SoPivot results cover 2,258 dialogues,
each rated three times by each final judge (6,774 ratings per judge). Two
draws in different non-Hard configurations return invalid action arguments
in all attempts; both configurations are scored on their two valid draws.
Means average ratings within dialogue, dialogues within configuration,
configurations within scene, and scenes equally.

The situated-only, general-only and full SoPivot rows in
Table~\ref{tab:component-judges} use the same frozen bank, scene--role
configurations, actor settings, generation protocol, judging and aggregation.
Each component variant disables the other memory level while retaining the
enabled level's retrieval and applicability rules, entry cap and
memory-token guard. Both judges score exactly the same dialogues.

\subsection{Assets and Reproducibility Scope}
The evaluated scenarios and dialogues are in English. The recorded runtime
uses Python 3.11.15, SOTOPIA 0.1.5, PettingZoo 1.25.0, LiteLLM 1.80.0 and
NumPy 2.4.4. SOTOPIA software is MIT-licensed. The source-profile snapshot
comes from \texttt{cmu-lti/sotopia-pi}, revision \texttt{e5834069}; its dataset
card specifies CC BY-SA 4.0.\footnote{\url{https://huggingface.co/datasets/cmu-lti/sotopia-pi}}
The official SOTOPIA dataset card lists a generic Creative Commons label
without specifying a variant; we do not infer dataset redistribution rights
from the software license.\footnote{\url{https://huggingface.co/datasets/cmu-lti/sotopia}}
These assets are used for simulated social-interaction research.

Inference and judging use remote APIs.

\subsection{Models and Deployment}
\label{app:models-deployment}
All names below refer to the API model identifiers recorded by the runs:
\texttt{qwen3.5-flash}, \texttt{qwen3.8-max}, and
\texttt{deepseek-v4-pro}.
Source actors and partners use Flash at temperature 0.7, a 500-token action
limit, disabled thinking, and a 20-turn episode limit. The selector and
the distinctness verifier use DeepSeek at temperature 0, and sibling
proposal uses DeepSeek at temperature 0.7 to diversify the alternatives;
each call has a 2,000-token output budget. Source outcomes use Max at temperature 0 with a
4,000-token scoring limit. Extraction uses DeepSeek at 0/2,000 tokens, with
Max checking source/context fidelity and action fidelity and up to two
quality repairs. Success eligibility is Goal $\ge7$, followed by maximum
unrounded mean and retention of all exact ties.

General-strategy admission uses three matched whole-episode replays per
condition and source setting, with Flash at temperature 0.7 and a 500-token
action limit, and Max at temperature 0 as the judge.

Target dialogue generation uses temperature 0, a 500-token action limit,
disabled thinking and 20 turns. The primary final judge uses Max, temperature 0, disabled
thinking and a 6,000-token limit. Only final judging receives the official
oracle fields, including goals needed to score each agent. It does not receive
method identity, retrieval metadata or source comparison evidence.

Student-visible queries and memory entries use
\texttt{text-embedding-v4} at 1,024 dimensions. For all three
SoPivot variants, situated retrieval checks the top ten candidates and
keeps at most five applicable entries; general retrieval checks all 34
entries and keeps at most three. DeepSeek performs applicability checking.
The single-level variants disable the other memory level without changing
these rules. All three use the same 12,000-token guard, and only the tested
agent sees the optional memory. Static baselines retain their native
content with at most three references and a 2,250-token cap.

\subsection{Aggregation}
Our evaluation matrix is 90 scenarios $\times$ five fixed ordered character
pairs $\times$ two focal roles. Hard has 14 scenarios (140 configurations);
the remaining 76 have 760 configurations. For each character pair, one
configuration assigns the focal agent to the first role and another assigns
it to the second, using the same partner model. Memory is provided only to
the focal agent, and reported scores correspond to that role.
With $J=3$ valid scores per dialogue, $R_c$ available dialogues per
configuration, and $C_s$ configurations in scene $s$,
\[
\begin{aligned}
\bar g_s&=\frac{1}{|C_s|}\sum_{c\in C_s}\frac{1}{R_c J}
\sum_{r=1}^{R_c}\sum_{j=1}^{J} g_{scrj},\\
\bar g&=\frac{1}{|S|}\sum_{s\in S}\bar g_s.
\end{aligned}
\]
Main-table coverage is 900 configurations per method, except PRINCIPLES
(899 under both judges), and Dialogue summary and MaTTS (899 under DeepSeek).
All Hard sets retain 140 configurations. Missing positions are excluded
from the means.

Goal is the primary outcome. Believability, relationship, knowledge, secret
keeping, social-rule adherence, and financial/material benefit are reported
in full in Appendix~\ref{app:full-results}.

\subsection{Environment Interface and Runtime Settings}
\label{app:environment-settings}
The two-agent environment uses round-robin actions. One turn is one
scheduled action by either participant, not a pair of utterances. Local
continuations resume the source turn counter; the 20-turn bound applies to
the full interaction, not an additional 20 turns after every branch.
Table~\ref{tab:runtime-settings} collects the shared settings.

\begin{table}[tbp]
\centering
\sopivottablestyle
\begin{tabular}{@{}>{\raggedright\arraybackslash}p{0.38\columnwidth}
>{\raggedright\arraybackslash}p{0.57\columnwidth}@{}}
\toprule
\textbf{Role or setting} & \textbf{Value} \\
\midrule
Environment & SOTOPIA 0.1.5; two agents; round-robin \\
Source / replay actors & Qwen3.5-Flash; temperature 0.7; 500 output tokens \\
Primary target / partner & Qwen3.5-Flash; temperature 0; 500 output tokens \\
Selection, verification, extraction & DeepSeek-V4-Pro; temperature 0; 2,000 output tokens \\
Sibling proposal & DeepSeek-V4-Pro; temperature 0.7; 2,000 output tokens \\
Content review & Qwen3.8-Max \\
Primary outcome judge & Qwen3.8-Max; temperature 0; 4,000 tokens for source branches, 6,000 for replay and target scoring \\
Second outcome judge & DeepSeek-V4-Pro; temperature 0; 6,000 output tokens \\
Thinking generation & Disabled for actors and outcome judges \\
Episode limit & 20 individual actions, including the restored prefix \\
Stale-action parameter & 2; stop after more than 2 consecutive in-turn \texttt{none} actions \\
Final ratings & Three calls per target dialogue; average the focal role's scores \\
\bottomrule
\end{tabular}
\caption{Runtime settings for source construction and primary evaluation.
Module-specific retrieval settings appear in
Appendix~\ref{app:models-deployment}.}
\label{tab:runtime-settings}
\end{table}

The action interface permits \texttt{speak}, \texttt{non-verbal communication},
\texttt{action}, \texttt{none}, and \texttt{leave}. These respectively produce
speech, a nonverbal signal, a described physical action, no action, or an exit;
described actions are textual events rather than physical-world execution.
The environment terminates at the turn limit, when fewer than two participants
remain active, or when the stale-action count exceeds two. Out-of-turn
placeholder actions are not added to the interaction history and do not
contribute to this count.

At reset, each actor receives the scenario, participant names, its own
rendered profile and goal; the separate partner profile and goal fields are
rendered as \texttt{Unknown}. The shared scenario itself may describe the
partner or relationship. Subsequent observations contain public interaction
events, the turn number, and available actions. Only the
focal agent receives retrieved memory. The retrospective outcome judge
receives the full evaluator history, including both goals and background
information; this oracle view is not passed to the actors.

\subsection{Score Dimensions and Interpretation}
\label{app:score-dimensions}
Table~\ref{tab:score-dimensions} defines the native SOTOPIA dimensions
\citep{zhou2024sotopia}. Individual ratings are integer-valued; reported
means need not be. Higher is better for all seven dimensions. Zero is the
maximum for secrecy and social-rule adherence, but denotes no change for
relationship and financial/material benefit.

\begin{table}[tbp]
\centering
\sopivottablestyle
\begin{tabular}{@{}>{\raggedright\arraybackslash}p{0.22\columnwidth}
c >{\raggedright\arraybackslash}p{0.49\columnwidth}@{}}
\toprule
\textbf{Dimension} & \textbf{Range} & \textbf{What is assessed?} \\
\midrule
Goal & $[0,10]$ & Achievement of the role's stated goal. \\
Relationship & $[-5,5]$ & Improvement or damage to relationships and social standing. \\
Knowledge & $[0,10]$ & Acquisition of new, relevant information. \\
Believability & $[0,10]$ & Natural behavior consistent with the character. \\
Secret & $[-10,0]$ & Disclosure of private information or secret intentions; zero means none disclosed. \\
Social rules & $[-10,0]$ & Violations of moral rules or laws; zero means none identified. \\
Financial / material & $[-5,5]$ & Gains or losses in material resources and opportunities. \\
\bottomrule
\end{tabular}
\caption{Seven dimensions and their native score ranges. The exact
judge instructions appear in Appendix~\ref{app:judge-rubrics}.}
\label{tab:score-dimensions}
\end{table}

The reported $\mathrm{Avg}=\frac{1}{7}\sum_{d=1}^{7}s_d$ is the equally
weighted arithmetic mean of the seven raw scores, without normalization or
shifting their origins. Goal is the primary endpoint and Avg is an auxiliary
summary.

\subsection{Reproduction Order}
\label{app:reproduction-order}
Code and accompanying research materials will be publicly released. The
algorithms, prompts, environment interface and settings in this appendix
document the protocol.

A reproduction first fixes the source/development allocation and the official
scenario--character--role configurations. It then generates source dialogues,
collects the sibling continuations, scores them, and performs selection and
reviewed extraction (Algorithm~\ref{alg:local-contrast}). Next it constructs
general candidates, executes paired source replay, and freezes the admitted
bank (Algorithm~\ref{alg:general-admission}). It then computes and freezes
the target retrieval bundles, generates each target dialogue, and records
three final ratings per judge. Aggregation joins by scenario, ordered
characters, focal role, method and dialogue identity, then follows the
scene-level procedure above.

\subsection{AgentSense Experimental Setup}
\label{app:agentsense-setup}
\paragraph{Benchmark and sample.}
AgentSense evaluates goal-directed social interaction and reasoning about
private information through role-play and post-dialogue questions
\citep{mou2025agentsense}. We use its two-person subset: 730 scenes from
146 templates, with five character variants per template. Each role is
in turn treated as the focal agent, giving 1,460 role configurations per
method. No memory, MaTTS and SoPivot use the same scene and role set.

\paragraph{Interaction and memory.}
Both participants use Qwen3.5-Flash with conversation temperature 1 and
128 output tokens per request. We retain the native prompts and a limit
of 15 total messages, including the initial greeting, with the same seeded
speaker schedule across methods. Each actor receives the public scene,
its own profile, goals and private information, and the observed dialogue.
The partner's hidden information and evaluation answer labels are excluded.
The memory methods transfer frozen source memories without learning from
AgentSense outcomes. Retrieval occurs once before the dialogue; the fixed
optional bundle is included only in the focal agent's dialogue requests.
SoPivot uses up to five situated and three general entries; MaTTS uses
up to three applicable entries from its retrieved candidates. The partner
receives no memory. Empty-memory configurations remain in the evaluation
and reuse the matched no-memory dialogue. Each nonempty-memory configuration
uses one dialogue; post-dialogue interviews receive no memory bundle.

\paragraph{Metrics and judging.}
All four columns in Table~\ref{tab:agentsense} are percentages, with larger
values indicating more positive judgments or correct answers.
\textbf{Goal} is the fraction of the focal agent's goals judged completed
by DeepSeek-V4-Pro from the dialogue, using the native goal-specific
questions. \textbf{Self} measures goal completion according to the focal
agent's own post-dialogue answers; \textbf{Other} uses the partner's answers
about the focal agent's goals.
\textbf{Info} is accuracy on the benchmark's multiple-choice questions
requiring inference of private information from the interaction, scored
against the supplied answers. External judging and the Flash interviews
use temperature 0 and a 128-token output limit. We retain the native
Yes/choice parsing and use one external judgment per goal.

\paragraph{Aggregation.}
Goal, Self and Other first average the native question judgments within
each goal and the goals within the focal role. We then average the two
focal-role configurations within each scene, the five variants within each
template, and the 146 templates equally. Info follows the native aggregation
on available questions; scenes without an Info question are not counted as
incorrect. The reported percentages are therefore template-weighted means.

\section{Baseline Adaptations and Component Controls}
\label{app:baselines}

\subsection{Source-Learning Adaptations}
All static methods start from the same 170 raw source-task pool, not the
subset that yields accepted SoPivot strategies. They retain their own
source filters and representations. DeepSeek is the common source-learning
model.

\paragraph{No memory and dialogue summaries.}
No memory uses the shared base actor prompt. The summary baseline reads each
complete student-visible source experience and writes a separate lesson;
146 entries survive construction checks. It does not receive local sibling
outcomes.

\paragraph{Direct advice.}
DeepSeek writes guidance from the source task's initial visible context with
an empty dialogue prefix and no source or target outcomes. This yields
138 accepted entries.

\paragraph{ReasoningBank (online).}
Our SOTOPIA adaptation \citep{ouyang2026reasoningbank} starts with 470
lessons grouped into 170 source experiences. At each task's
start, cosine retrieval selects one whole experience, including its lessons,
using the student's visible opening context. After the dialogue, the Flash
actor self-labels success or failure and extracts lessons for its next task;
final oracle grades are not used for learning. Ten isolated character-pair/role
streams each process 90 tasks, without sharing online memories.

The adaptation keeps whole-experience retrieval and online updates. Memory is
injected under the shared action-only JSON protocol; the seed bank comes
from the source experiences, the query uses the visible opening context, and
embeddings use 1,024-dimensional text-embedding-v4. The actor uses the
shared temperature 0 and 500-token action limit.

\paragraph{PRINCIPLES.}
The adaptation of \citet{kim-etal-2025-principles} constructs 199 principles
from the fixed 170 source tasks. It preserves ten sampled source-critic
judgments per trial, adjacent-trial mean comparisons, up to three backtrack
revisions, and the native conditional principle format. No SoPivot admission
or applicability filter is added. At inference, each turn retrieves three
principles by raw L2 distance and separately reinterprets them for the visible
history and own goal; the bank remains frozen.

Flash handles dialogue, planning, revisions and reinterpretation;
DeepSeek-V4-Pro extracts principles, and Max supplies source criticism and
final evaluation. The source critic maps visible goal status to four levels
($-1,-0.5,0.5,1$); this replaces the original domain-specific progress labels.
The embedding model is text-embedding-v4 (1,024 dimensions). Target dialogue
uses the common SOTOPIA stopping rules, temperature 0 and 500-token action
limit.

There are 899 valid target dialogues and 2,697 final grades; one non-Hard
configuration fails after action truncation and the standard format repair.
All 140 Hard configurations complete. Table~\ref{tab:main-results}
weights all 90 scenes equally, averaging the nine available configurations
in the incomplete scene.

\paragraph{MaTTS.}
The source-learning adaptation adds three
complete attempts per source task (510 dialogues) and compares the experiences,
yielding 659 entries. Its memory is frozen for target reuse.

\paragraph{ACE source replay.}
An independent reflector and curator process the 170 completed source
experiences once in a fixed order, incrementally adding, updating, merging
or removing playbook entries \citep{zhang2025ace}. The final playbook has
92 entries. Target deployment freezes the playbook and retrieves its native
entries.

\paragraph{Shared retrieval.}
Static baselines reuse compatible initial queries, recall ten candidates and
apply the applicability check before rendering complete native entries.
Selected memory is capped at three entries and 2,250 tokens. Online
ReasoningBank instead retrieves one whole experience without this
applicability filter, and its memory evolves across tasks.

\subsection{Removing Source Rollout Evidence}
\label{app:no-source-rollout}
This system ablation retains 335 source decision points and their three
candidate actions, but generates no source continuations and supplies no
continuation outcomes or scores to selection or extraction. DeepSeek-V4-Pro
chooses one action from the decision-time student-visible context, then
distills a situated strategy from that context and action. Grounding review
checks fidelity and information boundaries, rather than observed effectiveness.
The rebuilt bank contains 310 situated and 37 general strategies. The
treatment thus removes outcome evidence during memory construction while
retaining candidate comparison and two-level memory.

Evaluation covers the same 900 official scene--role configurations, including
140 Hard configurations. Retrieval allows up to five situated and three
general entries. There are 803
configurations with three generated dialogues each and 97 configurations
reusing one matched no-memory dialogue. Each dialogue receives three Max
grades. The completed run contains 2,409 new dialogues and 7,227 new grades.
Scores average judges and available dialogues within configuration, then
configurations within scene and scenes equally. Avg averages the seven raw
score dimensions. Table~\ref{tab:no-source-rollout} reports these means and
the full-system values from Table~\ref{tab:main-results}.

\section{Prompt Templates}
\label{app:prompts}
We reproduce the fixed instructions for selection, extraction, validation,
consolidation, and reuse. Named slots in angle brackets replace variable
scenario, dialogue, strategy, and identifier values. They are filled with
the indicated evidence, not passed literally to the model. Run settings
are specified in Appendix~\ref{app:evaluation}.
Quoted dialogue is data, not an instruction to the controller. Output
selection and source-admission rules are implemented in code
(Appendix~\ref{app:algorithms}), not delegated to a prompt.

\subsection{Retrospective decision-point selection}
The selector sees the completed public source dialogue and returns at most two original nodes. Reasons remain controller-side.
\sopivotpromptbox{Retrospective decision-point selection}{prompts/01_selection.txt}

\subsection{Sibling-action proposal}
The observed action is retained by the controller; the model proposes two alternatives. No controller-memory block is supplied here.
\sopivotpromptbox{Sibling-action proposal}{prompts/02_proposal.txt}

\subsection{Functional-distinctness verification}
A separate call checks every candidate pair. Outcome scores and winner prediction are outside this verifier.
\sopivotpromptbox{Functional-distinctness verification}{prompts/03_distinctness.txt}

\subsection{Base actor and natural continuation}
This is the native agent-mode template. The history is the acting character's observation history. Local comparison fixes one action, then resumes this policy; memory reuse adds the selected optional block only to the focal agent.
The action-list slot enumerates \texttt{none}, \texttt{speak},
\texttt{non-verbal communication}, \texttt{action}, and \texttt{leave};
their interface and termination rules are described in
Appendix~\ref{app:environment-settings}.
\sopivotpromptbox{Base actor and natural continuation}{prompts/13_actor.txt}

\subsection{Joint situated-strategy extraction}
The system instruction includes the source-fidelity rules. Variable evidence values are replaced by named slots; the output schema is retained.
Prompt~\ref{prompt:extraction-wide} follows in two full-width panels.
\refstepcounter{sopivotprompt}
\sopivotwidepromptpart{5}{Joint situated-strategy extraction (1/2)}{prompts/layout/04_extraction_part1.txt}{Extraction instructions and evidence boundaries (part 1 of 2).\label{prompt:extraction-wide}}{p}
\sopivotwidepromptpart{5}{Joint situated-strategy extraction (2/2)}{prompts/layout/04_extraction_part2.txt}{Source-fidelity rules, output schema, and user payload (part 2 of 2).}{p}

\FloatBarrier
\subsection{Isolated pre-action context review}
This review receives pre-action sources and proposed fields, excluding candidate actions and downstream trajectories. Exact supporting quotations are checked by code.
\sopivotpromptbox{Isolated pre-action context review}{prompts/05_context_review.txt}

\subsection{Selected-action fidelity review}
A separate review checks the selected move and the draft. The implementation allows at most two quality repairs, with the winner IDs fixed.
\sopivotpromptbox{Selected-action fidelity review}{prompts/06_action_review.txt}

\subsection{Bounded extraction repair}
Append this feedback to the original extraction messages. Each response must remain in the original schema and is reviewed again; at most two repairs are permitted.
\sopivotpromptbox{Bounded extraction repair}{prompts/14_repair.txt}

\subsection{Compatibility adjudication}
Semantic similarity proposes pairs; both compatibility criteria must pass. The rationale is an audit field.
\sopivotpromptbox{Compatibility adjudication}{prompts/07_merge.txt}

\subsection{General-strategy abstraction}
The supporting list contains all members of one approved group. Admission is determined by source replay, not by this language-model response.
\sopivotpromptbox{General-strategy abstraction}{prompts/08_general.txt}

\subsection{Episode-start applicability check}
The strict strategy verifier checks whether a reference is worth carrying. The action is withheld from this check; an optional conversational cue may be deferred. Empty failure scope does not add a failure-memory intervention.
Prompt~\ref{prompt:applicability-wide} follows in two full-width panels.
\refstepcounter{sopivotprompt}
\sopivotwidepromptpart{11}{Episode-start applicability check (1/2)}{prompts/layout/09_applicability_part1.txt}{Role, information boundaries, and input definitions (part 1 of 2).\label{prompt:applicability-wide}}{!b}
\sopivotwidepromptpart{11}{Episode-start applicability check (2/2)}{prompts/layout/09_applicability_part2.txt}{Decision rules, output schema, and payload fields (part 2 of 2).}{p}

\FloatBarrier
\subsection{Target-episode memory block}
The cards array contains the selected entries; general entries use general\_strategy in place of strategy\_label. This block remains visible to the tested agent throughout the episode; it is not sent to the partner.
\sopivotpromptbox{Target-episode memory block}{prompts/10_reuse.txt}

\subsection{General-memory source-replay block}
Source admission restarts complete source settings with general memory versus no memory.
\sopivotpromptbox{General-memory source-replay block}{prompts/11_source_replay.txt}

\subsection{Official outcome-evaluation wrapper}
The schema slot uses the official seven-dimension descriptions and ranges for both agents. Judge history includes the official oracle fields, not method labels or retrieval evidence.
\sopivotpromptbox{Official outcome-evaluation wrapper}{prompts/12_judge.txt}

\subsection{Complete Seven-Dimension Judge Rubric}
\label{app:judge-rubrics}
The following text reproduces the seven field descriptions embedded in the
official \texttt{SotopiaDimensions} schema, rather than a paraphrase of the
scoring rules. Dimension names are headings for presentation; the wrapper
above receives the generated JSON schema containing these descriptions.
Every dimension returns an object with a justification string under
\texttt{reasoning} and an integer under \texttt{score}. The outer structure is
\texttt{evaluations}, containing exactly \texttt{agent\_1} and
\texttt{agent\_2}; each contains all seven dimension keys shown below.
The focal role determines which agent's scores enter the reported result;
it is not always \texttt{agent\_1}. Repeated judging does not change this
mapping or the rubric. Score ranges and their interpretation are summarized
in Appendix~\ref{app:score-dimensions}.
\sopivotpromptbox{Official seven-dimension rubric}{prompts/15_judge_rubrics.txt}

\FloatBarrier

\section{Illustrative Strategy Example}
\label{app:illustration}

Consider a constructed allocation dialogue: the student wants the books,
and the partner explicitly says, ``I care most about the hats.'' This public
utterance makes the stated preference observable. The example illustrates
the strategy representation and its conditional use.

\paragraph{Situated strategy.}
\emph{Visible condition:} the parties have stated different priorities over
divisible items and can propose an allocation.
\emph{Strategy move:} state the high-priority claim and offer the items the
counterpart publicly prioritized as a reciprocal trade.
\emph{Expected interaction shift:} move from ambiguous bargaining to a
concrete proposal addressing the stated priorities.

An illustrative action asks for the books while offering the hats. It does
not justify guessing an unspoken preference or assuming agreement. A general
version removes item names but retains reciprocal allocation. Reuse still
requires visible differing priorities and a feasible trade in the new interaction.

\raggedbottom
\section{Full Official-Set Results}
\label{app:full-results}
This section expands the main comparison along three questions: which outcome
dimensions change, whether the aggregate pattern also holds on Hard scenes,
and how often each system supplies memory. All comparisons use the
predefined subsets.

\subsection{Overall Outcome Profile}
\begin{table*}[!t]
\centering
\sopivottablestyle
\begin{tabular*}{\linewidth}{@{\extracolsep{\fill}}lrrrrrrr@{}}
\toprule
\textbf{Method} & \textbf{Bel $\uparrow$} & \textbf{Rel $\uparrow$} & \textbf{Kno $\uparrow$} & \textbf{Sec $\uparrow$} & \textbf{Soc $\uparrow$} & \textbf{Fin $\uparrow$} & \textbf{Goal $\uparrow$} \\
\midrule
No memory & 7.374 & 0.895 & 5.818 & -1.359 & -0.911 & 0.730 & 6.624 \\
\addlinespace[3pt]
Dialogue summary & 7.357 & 0.984 & 5.843 & -1.267 & -0.808 & 0.701 & 6.595 \\
Direct advice & 7.336 & 1.006 & 5.877 & -1.265 & -0.815 & 0.714 & 6.657 \\
ReasoningBank & 7.283 & 1.276 & 6.030 & -1.146 & -0.574 & 0.826 & 6.704 \\
MaTTS & \textbf{7.424} & 1.419 & \underline{6.210} & \underline{-0.876} & -0.558 & \textbf{0.907} & 6.942 \\
ACE & \underline{7.387} & 1.370 & 6.004 & -1.087 & -0.603 & 0.872 & 6.883 \\
PRINCIPLES$^{\dagger}$ & 4.131 & 0.768 & 5.131 & \textbf{-0.401} & \textbf{-0.371} & 0.783 & 5.727 \\
\midrule
Situated only & 7.323 & 1.094 & 5.962 & -1.246 & -0.764 & 0.743 & 6.891 \\
General only & 7.374 & \underline{1.517} & 6.148 & -1.083 & -0.498 & 0.859 & \underline{7.131} \\
\addlinespace[2pt]
\textbf{SoPivot (Ours)} & 7.354 & \textbf{1.629} & \textbf{6.227} & -1.067 & \underline{-0.435} & \underline{0.890} & \textbf{7.220} \\
\bottomrule
\end{tabular*}
\caption{Seven-dimensional means on SOTOPIA-All (90 scenes). Higher is better: Bel/Kno/Goal [0,10], Rel/Fin [-5,5], Sec/Soc [-10,0]. \textbf{Bold}/\underline{underlining} mark the best/second-best displayed values, including ties. SoPivot uses the source-replay-admitted 34-general bank. PRINCIPLES$^{\dagger}$ averages nine configurations in one incomplete scene (899/900 total).}
\label{tab:dimensions-all90}
\end{table*}

\begin{table*}[!t]
\centering
\sopivottablestyle
\begin{tabular*}{0.85\linewidth}{@{\extracolsep{\fill}}lrr@{}}
\toprule
\textbf{Compared with} & \textbf{All 90 $\uparrow$} & \textbf{Hard 14 $\uparrow$} \\
\midrule
No memory & $+0.596$ & $+0.339$ \\
\addlinespace[3pt]
Dialogue summary & $+0.625$ & $+0.297$ \\
Direct advice & $+0.563$ & $+0.494$ \\
ReasoningBank & $+0.516$ & $+0.406$ \\
MaTTS & $+0.278$ & $-0.053$ \\
ACE & $+0.337$ & $+0.225$ \\
PRINCIPLES$^{\dagger}$ & $+1.493$ & $+1.156$ \\
\midrule
Situated only & $+0.329$ & $+0.262$ \\
General only & $+0.089$ & $+0.099$ \\
\bottomrule
\end{tabular*}
\caption{SoPivot Goal minus each comparator, calculated from the Max means in Tables~\ref{tab:main-results} and~\ref{tab:component-judges}. }
\label{tab:main-intervals}
\end{table*}

Table~\ref{tab:dimensions-all90} reports scene-equal means on the same 90
scenes. Bel, Rel, Kno, Sec, Soc and Fin denote believability, relationship,
knowledge, secret keeping, social-rule adherence and financial/material
benefit. Higher is better, but the dimensions have different ranges; a raw
change of one point is not interchangeable across them.

Relative to no memory, SoPivot has All Goal 7.220 versus 6.624,
Relationship 1.629 versus 0.895, and Avg 3.117 versus 2.739.
The corresponding Goal and Avg gains are 0.596 and 0.378.
Table~\ref{tab:main-intervals} reports the Goal differences between
SoPivot and each comparator.

Under Max on All, Relationship and Knowledge improve over no memory by
0.734 and 0.409, respectively, while Believability stays close to no memory
(7.354 versus 7.374). The gains are thus concentrated in interaction
outcomes. MaTTS has higher Believability and Financial scores, and
PRINCIPLES has higher Secret and Social-rule scores.

\FloatBarrier
\subsection{Performance on Hard Scenarios}
\begin{table*}[!t]
\centering
\sopivottablestyle
\begin{tabular*}{\linewidth}{@{\extracolsep{\fill}}lrrrrrrr@{}}
\toprule
\textbf{Method} & \textbf{Bel $\uparrow$} & \textbf{Rel $\uparrow$} & \textbf{Kno $\uparrow$} & \textbf{Sec $\uparrow$} & \textbf{Soc $\uparrow$} & \textbf{Fin $\uparrow$} & \textbf{Goal $\uparrow$} \\
\midrule
No memory & 7.240 & -1.307 & 5.117 & -1.186 & -1.898 & 0.036 & 5.060 \\
\addlinespace[3pt]
Dialogue summary & 7.174 & -1.148 & 5.174 & -0.936 & -1.702 & 0.112 & 5.102 \\
Direct advice & 7.310 & -1.131 & 5.250 & -0.943 & -1.569 & -0.050 & 4.905 \\
ReasoningBank & 7.402 & -0.636 & 5.398 & -0.833 & -0.933 & 0.250 & 4.993 \\
MaTTS & \underline{7.524} & -0.512 & 5.476 & -0.790 & -1.150 & \textbf{0.500} & \textbf{5.452} \\
ACE & \textbf{7.543} & -0.629 & 5.329 & -0.810 & -1.331 & 0.355 & 5.174 \\
PRINCIPLES$^{\dagger}$ & 4.826 & \textbf{0.129} & 4.971 & \textbf{-0.340} & \textbf{-0.583} & -0.079 & 4.243 \\
\midrule
Situated only & 7.113 & -1.052 & 5.279 & -0.987 & -1.575 & 0.026 & 5.137 \\
General only & 7.510 & -0.330 & \underline{5.604} & -0.599 & -0.932 & 0.315 & 5.300 \\
\addlinespace[2pt]
\textbf{SoPivot (Ours)} & 7.435 & \underline{-0.211} & \textbf{5.638} & \underline{-0.525} & \underline{-0.845} & \underline{0.371} & \underline{5.399} \\
\bottomrule
\end{tabular*}
\caption{Seven-dimensional means on SOTOPIA-Hard (14 scenes; 140 configurations per method). Higher is better: Bel/Kno/Goal [0,10], Rel/Fin [-5,5], Sec/Soc [-10,0]. \textbf{Bold}/\underline{underlining} mark the best/second-best displayed values, including ties. SoPivot uses the source-replay-admitted 34-general bank.}
\label{tab:dimensions-hard14}
\end{table*}

Table~\ref{tab:dimensions-hard14} reports the 14 Hard scenes. SoPivot Goal
is 5.399, versus 5.060 without memory and 5.452 for MaTTS.
Its Goal gain over no memory is 0.339. Hard Avg is 2.466,
versus 1.866 without memory, a gain of 0.600.

\subsection{Memory Coverage and the Component Comparison}
\label{app:component-results}
\begin{table*}[!t]
\centering
\sopivottablestyle
\begin{tabular*}{\linewidth}{@{\extracolsep{\fill}}lrrrrrrrr@{}}
\toprule
 & \multicolumn{4}{c}{\textbf{Qwen3.8-Max}} & \multicolumn{4}{c}{\textbf{DeepSeek-V4-Pro}} \\
\cmidrule(lr){2-5}\cmidrule(l){6-9}
 & \multicolumn{2}{c}{\textbf{All (90)}} & \multicolumn{2}{c}{\textbf{Hard (14)}} & \multicolumn{2}{c}{\textbf{All (90)}} & \multicolumn{2}{c}{\textbf{Hard (14)}} \\
\cmidrule(lr){2-3}\cmidrule(lr){4-5}\cmidrule(lr){6-7}\cmidrule(l){8-9}
\textbf{Method} & \textbf{Goal $\uparrow$} & \textbf{Avg $\uparrow$} & \textbf{Goal $\uparrow$} & \textbf{Avg $\uparrow$} & \textbf{Goal $\uparrow$} & \textbf{Avg $\uparrow$} & \textbf{Goal $\uparrow$} & \textbf{Avg $\uparrow$} \\
\midrule
Situated only & 6.891 & 2.858 & 5.137 & 1.992 & 6.363 & 2.773 & 4.421 & 1.878 \\
General only & \underline{7.131} & \underline{3.064} & \underline{5.300} & \underline{2.410} & \underline{6.513} & \underline{2.896} & \underline{4.533} & \underline{2.090} \\
SoPivot (Ours) & \textbf{7.220} & \textbf{3.117} & \textbf{5.399} & \textbf{2.466} & \textbf{6.822} & \textbf{2.995} & \textbf{4.986} & \textbf{2.240} \\
\bottomrule
\end{tabular*}
\caption{Memory-component results under both judges, using the same frozen bank and deployment rules. Goal is the primary metric; Avg averages seven raw dimension scores. Bold/underlining mark the best/second-best results (ties included). Evaluation details are in Appendix~\ref{app:evaluation}.}
\label{tab:component-judges}
\end{table*}

\begin{table*}[!t]
\centering
\sopivottablestyle
\begin{tabular*}{0.85\linewidth}{@{\extracolsep{\fill}}lrr@{}}
\toprule
\textbf{Method} & \textbf{All} & \textbf{Hard} \\
\midrule
Dialogue summary & 198/900 & 43/140 \\
Direct advice & 177/900 & 26/140 \\
ReasoningBank & 900/900 & 140/140 \\
MaTTS & 727/900 & 110/140 \\
ACE & 532/900 & 99/140 \\
\midrule
\addlinespace[2pt]
\textbf{SoPivot (Ours)} & 751/900 & 134/140 \\
\bottomrule
\end{tabular*}
\caption{Configurations receiving nonempty memory. ReasoningBank uses online updates and whole-experience top-1 retrieval; the remaining rows use static memory. Empty-memory configurations remain in evaluation and use the matched no-memory dialogues.}
\label{tab:coverage}
\end{table*}

The component comparison fixes the bank, target
configurations, retrieval rules, token guard, generation protocol and judge.
It changes which memory levels are enabled. Under Max, situated-only,
general-only and full SoPivot score 6.891, 7.131 and 7.220 on All,
and 5.137, 5.300 and 5.399 on Hard. Full SoPivot also has higher Avg
than either single-level variant on both subsets (Table~\ref{tab:component-judges}).

Table~\ref{tab:coverage} reports full-system and baseline memory coverage.
Full SoPivot supplies memory to 751/900 configurations (134/140 Hard).
Across static baselines, coverage ranges from 177/900 for direct advice to
727/900 for MaTTS. Online ReasoningBank supplies memory to all 900
configurations: 395 retrieve a warm source experience and 505 an experience
from an earlier target task in the same stream. Empty-memory configurations
remain part of each system's evaluation.

\FloatBarrier
\section{Guidance at Source Decision Points}
\label{app:source-diagnostic}
\begin{table*}[!t]
\centering
\sopivottablestyle
\begin{tabular*}{\linewidth}{@{\extracolsep{\fill}}lrrrrrrr@{}}
\toprule
\textbf{Condition} & \textbf{Bel $\uparrow$} & \textbf{Rel $\uparrow$} & \textbf{Kno $\uparrow$} & \textbf{Sec $\uparrow$} & \textbf{Soc $\uparrow$} & \textbf{Fin $\uparrow$} & \textbf{Goal $\uparrow$} \\
\midrule
No guidance & 7.759 & 2.321 & 5.963 & -0.778 & -0.204 & 1.019 & 8.475 \\
Situated & \textbf{8.630} & \textbf{3.395} & \textbf{6.735} & \textbf{-0.235} & \textbf{-0.006} & \textbf{1.167} & \textbf{9.420} \\
General & \underline{8.241} & \underline{3.074} & \underline{6.494} & \underline{-0.321} & \underline{-0.037} & \underline{1.154} & \underline{9.185} \\
\bottomrule
\end{tabular*}
\caption{Guidance at source decision points: seven-dimensional means with the nine source scenes weighted equally. Each condition uses three generated continuations per node, each scored three times. \textbf{Bold}/\underline{underlining} mark the best/second-best values.}
\label{tab:source-dimensions}
\end{table*}

The main experiments evaluate memory through episode-start retrieval in new
scenarios. Here we ask a more local question: at the decision points where
strategies were discovered, does supplying a strategy improve the outcome,
and does a general strategy retain the benefit of the situated strategies it
abstracts?

We test five general candidates produced by consolidation together with
their 12 situated members, covering 11 source decision points in nine
scenes. At each decision point, the focal agent receives no guidance, the
situated strategy, or the general strategy for one action, after which the
guidance is removed and the dialogue continues naturally. Each condition
uses three fresh dialogues, each judged three times. Both actors are Flash
at temperature 0 with a 500-token action limit, and Max supplies the scores.
The situated condition generates new continuations and does not reuse the
continuations from strategy discovery.

Table~\ref{tab:source-dimensions} weights the nine scenes equally. Situated
guidance raises Goal from 8.475 to 9.420, and general guidance raises it to
9.185. Both forms of guidance also improve the other six dimensions over no
guidance, and the general strategy lies between no guidance and the situated
strategy on all seven. The scene-paired difference in Goal between general
and situated guidance is $-0.235$, with a 95\% bootstrap interval of
[$-0.809$, 0.160]. General strategies therefore retain most of the local
benefit of their situated sources, which is consistent with the
complementary roles of the two memory levels in Section~\ref{sec:ablation}.

\FloatBarrier
\section{Judge Sensitivity Analysis}
\label{app:robustness}
\begin{table*}[!t]
\centering
\sopivottablestyle
\begin{tabular*}{\linewidth}{@{\extracolsep{\fill}}lrrrrrrrr@{}}
\toprule
\textbf{Judge / Set} & \textbf{Bel $\uparrow$} & \textbf{Rel $\uparrow$} & \textbf{Kno $\uparrow$} & \textbf{Sec $\uparrow$} & \textbf{Soc $\uparrow$} & \textbf{Fin $\uparrow$} & \textbf{Goal $\uparrow$} & \textbf{Avg $\uparrow$} \\
\midrule
Max / All & 7.354 & 1.629 & 6.227 & -1.067 & -0.435 & 0.890 & 7.220 & 3.117 \\
DeepSeek / All & 8.466 & 1.373 & 4.733 & -0.565 & -0.310 & 0.448 & 6.822 & 2.995 \\
\midrule
Max / Hard & 7.435 & -0.211 & 5.638 & -0.525 & -0.845 & 0.371 & 5.399 & 2.466 \\
DeepSeek / Hard & 8.231 & -0.302 & 3.885 & -0.344 & -0.673 & -0.102 & 4.986 & 2.240 \\
\bottomrule
\end{tabular*}
\caption{SoPivot under two judges on exactly the same dialogues. All/Hard contain 900/140 configurations. Three grades are averaged per dialogue, followed by available dialogues per configuration and equal scene weights. Avg is the mean of the seven raw dimensions. Goal and Avg match Table~\ref{tab:main-results}.}
\label{tab:current-judge-scores}
\end{table*}

Both judges evaluate the same 2,258 SoPivot dialogues from the 900 configurations, with 6,774 ratings per judge (Appendix~\ref{app:expanded-bank}). Table~\ref{tab:current-judge-scores} gives both seven-dimensional profiles. The DeepSeek Goal gain over no memory is 0.481 on All and 0.593 on Hard; the corresponding Avg gains are 0.293 and 0.522.

\end{document}